# Iterated Belief Change Due to Actions and Observations

**Aaron Hunter**　　　　　　　　　　　　　　　　　　　　　　　HUNTER@CS.SFU.CA
*British Columbia Institute of Technology*
*Burnaby, BC, Canada*
**James P. Delgrande**　　　　　　　　　　　　　　　　　　　　JIM@CS.SFU.CA
*Simon Fraser University*
*Burnaby, BC, Canada*

## Abstract

In action domains where agents may have erroneous beliefs, reasoning about the effects of actions involves reasoning about belief change. In this paper, we use a transition system approach to reason about the evolution of an agent's beliefs as actions are executed. Some actions cause an agent to perform belief revision while others cause an agent to perform belief update, but the interaction between revision and update can be non-elementary. We present a set of rationality properties describing the interaction between revision and update, and we introduce a new class of belief change operators for reasoning about alternating sequences of revisions and updates. Our belief change operators can be characterized in terms of a natural shifting operation on total pre-orderings over interpretations. We compare our approach with related work on iterated belief change due to action, and we conclude with some directions for future research.

## 1. Introduction

We are interested in modeling the belief change that is caused by executing an alternating sequence of actions and observations. Roughly, agents perform *belief update* following actions and agents perform *belief revision* following observations. However, to date there has been little explicit consideration of the iterated belief change that results when an observation follows a sequence of actions. In this paper, we address belief change in the context of sequences of the following form.

$$(InitialBeliefs) \cdot (Action) \cdot (Observation) \cdots (Action) \cdot (Observation) \qquad (1)$$

We assume that the effects of actions are completely specified and infallible. We illustrate that the belief change caused by this kind of alternating sequence can not simply be determined through a straightforward iteration of updates and revisions. The main issue is that there are plausible examples where an observation should lead an agent to revise the initial belief state, rather than the current belief state. Our goal is to provide a general methodology for computing iterated belief change due to alternating sequences of actions and observations.

### 1.1 Contributions to Existing Research

This paper is an extension of our previous work (Hunter & Delgrande, 2005). Our focus is on action domains involving a single agent that can execute actions and make observations. It is assumed that every change in the state of the world is due to the execution of an action, and it is assumed that the set of available actions is given. In this context, there are two possible explanations for an erroneous belief. First, an erroneous belief can be due to an incorrect initial belief. Second, an erroneous belief can be due to the execution of a hidden exogenous action. We focus primarily on the first case, since we are explicitly concerned with a single agent scenario. We briefly describe the contributions that we make in the area of belief change caused by actions.

One contribution is that we explicitly specify precise properties that should hold whenever an action is followed by an observation. We state the properties in the style of the AGM postulates for belief revision (Alchourrón, Gärdenfors, & Makinson, 1985), and we argue that the properties should hold in any action domain involving a single agent with perfect knowledge of the actions executed. The properties of iterated belief change that we specify are a natural generalization of the AGM postulates; as such, they can easily be justified for action domains involving a given AGM operator. We show that there are simple examples where it is clear that the action history plays a role in the interpretation of an observation. Therefore, it is necessary to formalize the role of the action history in determining the appropriate belief change. However, this problem has not been explicitly addressed in related work. To the best of our knowledge, our work is the first attempt at formally specifying any high-level interaction between belief update and belief revision caused by actions.

As a second contribution, we give a specific methodology for combining a belief update operator and a belief revision operator in a single formalism. In particular, we define a new class of belief change operators, called *belief evolution* operators. A belief evolution operator takes two arguments: a set of states and an alternating sequence of actions and observations. Each belief evolution operator ∘ is defined with respect to a fixed update operator ⋄ and a fixed AGM revision operator ∗. Informally, we have the following correspondence

$$\kappa \circ \langle A_1, \alpha_1, \ldots, A_n, \alpha_n \rangle \approx \kappa \diamond A_1 * \alpha_1 \diamond \cdots \diamond A_n * \alpha_n.$$

The basic idea is simply to translate all observations into conditions on the initial beliefs. In this manner, we can define an iterated belief change operator that respects our interaction properties and handles example problems appropriately.

Formally, we demonstrate that our belief evolution operators can be characterized by a natural "shifting" on the underlying AGM revision operator. In this sense, we can view iterated belief change as a modified form of revision. From a more general perspective, the belief evolution methodology is useful for combining any action formalism with an AGM revision operator. Hence, we can view belief evolution as an improved methodology for adding a revision operator to an action formalism.

The third contribution of this work is that we provide a mechanism for evaluating the performance of existing epistemic action formalisms with regards to iterated belief change. It has been common in past work to extend an existing action formalism by simply adding a formal representation of knowledge or belief. This has been done, for example, in the action language $\mathcal{A}$ (Lobo, Mendez, & Taylor, 2001; Son & Baral, 2001) as well as the Situation Calculus (SitCalc) (Shapiro, Pagnucco, Lesperance, & Levesque, 2000). It is easy to see that the extensions of $\mathcal{A}$ fail to satisfy our interaction properties, and the sensing actions in the extended languages do not provide an appropriate model of iterated belief change. On the other hand, we suggest that the SitCalc does consider the action history appropriately. In general, we argue that simply extending an action formalism with a belief revision operator is not sufficient. We show that such approaches either lack the formal machinery required for reasoning about iterated belief change, or they make substantive implicit assumptions. Our work illustrates the role that an action formalism plays in reasoning about belief change. It also becomes clear that additional assumptions must be made in order to reason about iterated belief change due to actions and observations. By making the role of the action formalism salient, we can better evaluate the suitability of existing formalisms for particular applications.

## 2. Preliminaries

In this section, we introduce preliminary notation and definitions related to reasoning about action and reasoning about belief change. We also introduce a motivating example that will be used throughout the paper.

### 2.1 Motivating Example

We introduce Moore's litmus paper problem (Moore, 1985) as a motivating example. In this problem, there is a beaker containing either an acid or a base, and there is an agent holding a piece of white litmus paper that can be dipped into the beaker to determine the contents. The litmus paper will turn red if it is placed in an acid and it will turn blue if it is placed in a base. The problem is to provide a formal model of the belief change that occurs when an agent uses the litmus paper to test the contents of the beaker.

Intuitively, the litmus paper problem seems to require an agent to revise the initial beliefs in response to an observation at a later point in time. For example, suppose that the agent dips the paper and then sees that the paper turns red. This observation not only causes the agent to believe the beaker contains an acid *now*, but it also causes the agent to believe that the beaker contained an acid *before dipping*. We refer to this process as a *prior revision*, since the agent revises their beliefs at a prior point in time. This kind of phenomenon is not explicitly discussed in many formalisms for reasoning about belief change caused by action. We will return to this problem periodically as we introduce our formal approach to belief change.

### 2.2 Basic Notation and Terminology

We assume a *propositional signature* composed of a finite set of *atomic propositional symbols*. We use the primitive propositional connectives $\{\neg, \rightarrow\}$, where $\neg$ denotes classical negation and $\rightarrow$ denotes implication. Conjunction, disjunction and equivalence are defined in the usual manner, and they are denoted by $\wedge$, $\vee$ and $\equiv$, respectively. A *formula* is a propositional combination of atomic symbols. A *literal* is either an atomic propositional symbol, or an atomic propositional symbol preceded by the negation symbol. Let *Lits* denote the set of all literals over the fixed signature.

An *interpretation* of a propositional signature $\mathbf{P}$ is a function that assigns every atomic symbol a truth value. The set of all interpretations over $\mathbf{P}$ is denoted by $2^{\mathbf{P}}$. The satisfaction relation $I \models \phi$ is defined for formula $\phi$ by the usual recursive definition. For any formula $\phi$, we define $|\phi|$ to be the set of all interpretations $I$ such that $I \models \phi$, and we say that $\phi$ is satisfiable if and only if $|\phi| \neq \emptyset$.

A *belief state* is a subset of $2^{\mathbf{P}}$. We can think of a belief state as expressing a proposition. Informally, an agent with belief state $\kappa$ believes that the actual world is represented by one of the interpretations in $\kappa$. An *observation* is also a set of interpretations. The intuition is that the observation $\alpha$ provides evidence that the actual world is in the set $\alpha$. In order to maintain a superficial distinction, we will use the Greek letter $\alpha$ will range over observations and the Greek letter $\kappa$ to range over belief states, with possible subscripts in each case.

## 2.3 Transition Systems

An *action signature* is a pair $\langle \mathbf{A}, \mathbf{F} \rangle$ where $\mathbf{A}, \mathbf{F}$ are non-empty sets of symbols. We call $\mathbf{A}$ the set of *action symbols*, and we call $\mathbf{F}$ the set of *fluent symbols*. Formally, the fluent symbols in $\mathbf{F}$ are propositional variables. The action symbols in $\mathbf{A}$ denote the actions that an agent may perform. The effects of actions can be specified by a transition system.

**Definition 1** *A transition system $T$ for an action signature $\sigma = \langle \mathbf{A}, \mathbf{F} \rangle$ is a pair $\langle S, R \rangle$ where*

1. *$S$ is a set of propositional interpretations of $\mathbf{F}$,*

2. *$R \subseteq S \times \mathbf{A} \times S$.*

The set $S$ is called the set of *states* and $R$ is the *transition relation*. If $(s, A, s') \in R$, then we think of the state $s'$ as a possible resulting state that could occur if the action $A$ is executed in state $s$. If there is exactly one possible resulting state $s'$ when $A$ is executed in $s$ for all $s \in S$, then we say $T$ is *deterministic*. We refer to fluent symbols as being *true* or *false* if they are assigned the values $t$ or $f$, respectively. Transition systems can be visualized as directed graphs, where each node is labeled with a state and each edge is labeled with an element of $\mathbf{A}$. In terms of notation, the uppercase letter $A$, possibly subscripted, will range over actions. We use the notation $\bar{A}$ to denote a finite sequence of action symbols of indeterminate length. Also, given any sequence of actions $\bar{A} = \langle A_1, \ldots, A_n \rangle$, we write $s \leadsto_{\bar{A}} s'$ to indicate that there is a path from $s$ to $s'$ that follows the edges labeled by the actions $A_1, \ldots, A_n$.

It is useful to introduce the symbol $\lambda$ to denote the *null* action that never changes the state of the world. This will be used periodically in formal results. Also, for technical reasons, we assume throughout this paper that every action is executable in every state. If the transition system does not specify the effects of a particular action in a particular state, we assume that the state does not change when that action is executed. This is tantamount to adding self loops for every action at every state where no transition is given.

**Example** The litmus paper problem can be represented with the action signature

$$\langle \{dip\}, \{Red, Blue, Acid\} \rangle.$$

Intuitively, the fluent symbols *Red* and *Blue* represent the colour of the litmus paper, and the fluent symbol *Acid* indicates whether the beaker contains an acid or not. The only action available is to dip the litmus paper in the beaker

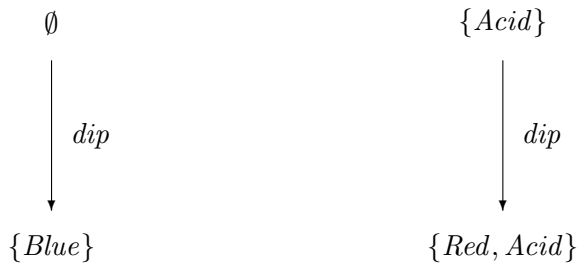

Figure 1: Litmus Test

In the interest of readability, we will adopt a notational convention in our discussion of the litmus paper problem. In particular, we will let a set $V$ of propositional fluent symbols stand for the *interpretation* where where the symbols in $V$ are assigned the value *true* and all other symbols are assigned the value *false*. Hence, the set $\{Red\}$ will be used to denote the interpretation $I$ where $I(Red) = true$, $I(Blue) = false$ and $I(Acid) = false$. This is a standard representation in the literature on reasoning about action effects. We stress, however, that we are not defining states in this manner; we are simply using this convention in the litmus paper example because it facilitates the specification of an interpretation over a small set of fluent symbols. To be clear, throughout this paper, states are actually *defined* in terms of interpretations over sets of propositional variables.

The effects of dipping in the litmus paper problem are given by the transition system in Figure 1. Note that we have not included all possible states in the figure; we have included only those states that change when an action is executed. We assume that the state remains unchanged when a dipping action is performed in any of the states omitted from the figure.

## 2.4 Belief Update

*Belief update* is the belief change that occurs when new information is acquired regarding a change in the state of the world. One standard approach to belief update is the Katsuno and Mendelzon approach (1991), which describes belief update in terms of a set of rationality postulates. These postulates are typically referred to as the KM postulates, and they have been analyzed, reformulated, and criticized in several subsequent papers (Boutilier, 1995; Peppas, Nayak, Pagnucco, Foo, Kwok, & Prokopenko, 1996; Lang, 2006). Much of this discussion has focused on the distinction between belief update and belief revision, which we introduce in the next section.

In this paper, we adopt an approach to belief update in which beliefs are updated by an *action* rather than a *formula*. Intuitively, after executing an action $A$, an agent updates its belief state by projecting every state $s$ to the state $s'$ that would result if the action $A$ was executed in the state $s$.

**Definition 2** *Let $T = \langle S, R \rangle$ be a transition system. The update function $\diamond : 2^S \times \mathbf{A} \to 2^S$ is defined as follows*

$$\kappa \diamond A = \{s' \mid (s, A, s') \in R \text{ for some } s \in \kappa\}.$$

This operation is actually a form of *action progression*; it has been argued elsewhere that the standard account of belief update can be understood to be a special case of this kind of progression (Lang, 2006). The advantage of our approach is that it provides a simple representation of the belief change that occurs following an action with conditional effects.

**Example** In the litmus paper problem, the agent believes the litmus paper is white, but it is not known whether the beaker contains an acid or a base. Hence, the initial belief state $\kappa$ consists of two interpretations that can be specified as follows:

$$\kappa = \{\emptyset, \{Acid\}\}.$$

After executing the *dip* action, the new belief state $\kappa \diamond dip$ consists of all possible outcomes of the *dip* action. Hence, the new belief state contains two interpretations. In the first, *Red* is true while the rest of the fluents are false. In the second interpretation, *Blue* and *Acid* are true, while the rest of the fluents are false. Thus, following a *dip* action, the agent believes that either the liquid is a base and the litmus paper is red or the liquid is an acid and the litmus paper is blue. To determine which outcome has occurred, the agent must observe the actual color of the paper.

## 2.5 Belief Revision

The term *belief revision* refers to the process in which an agent incorporates new information with some prior beliefs. In this section, we briefly sketch the most influential approach to belief revision: the AGM approach of Alchourrón, Gärdenfors and Makinson (1985). The AGM approach to belief revision does not provide a specific recipe for revision. Instead, a set of rationality postulates is given and any belief change operator that satisfies the postulates is called an AGM belief revision operator.

Let $\mathbf{F}$ be a propositional signature. A *belief set* is a deductively closed set of formulas over $\mathbf{F}$. Let $+$ denote the so-called *belief expansion operator*, which is defined by setting $K + \phi$ to be the deductive closure of $K \cup \{\phi\}$. Let $*$ be a function that maps a belief set and a formula to a new belief set. We say that $*$ is an AGM belief revision operator if it satisfies the following postulates for every $K$ and $\phi$.

[AGM1] $K * \phi$ is deductively closed

[AGM2] $\phi \in K * \phi$

[AGM3] $K * \phi \subseteq K + \phi$

[AGM4] If $\neg \phi \notin K$, then $K + \phi \subseteq K * \phi$

[AGM5] $K * \phi = \mathcal{L}$ iff $\models \neg \phi$

[AGM6] If $\models \phi \equiv \psi$, then $K * \phi = K * \psi$

[AGM7] $K * (\phi \wedge \psi) \subseteq (K * \phi) + \psi$

[AGM8] If $\neg \psi \notin K * \phi$, then $(K * \phi) + \psi \subseteq K * (\phi \wedge \psi)$

The main intuition behind the AGM postulates is that the new information given by $\phi$ must be incorporated, along with "as much of $K$" as consistently possible. The AGM postulates provide a simple set of conditions that are intuitively plausible as restrictions on belief revision operators. Moreover, the postulates completely determine a specific semantics for revision in terms of pre-orderings over interpretations (Katsuno & Mendelzon, 1992). We defer discussion of this semantics to §5, where we describe it in the context of a new belief change operator.

Note that we have presented the AGM postulates in the traditional setting, where beliefs and observations are given as sets of propositional formulae. In contrast, we represent beliefs and observations as sets of interpretations. However, since we work with a finite language, it is easy to translate between the two approaches and we will provide such translations when required.

**Example**  In the litmus paper problem, suppose that the paper turns red after dipping. We need to represent this observation in a suitable manner for revision. We stated in §2.2 that an observation is a set of interpretations. Informally, an observation is the set of interepretations that should be considered the most plausible after the observation. In the litmus paper example, the observation that the paper is red is represented by the set of all interpretations where $Red$ is true.

We need to revise the current beliefs by an observation that represents the information. Recall that the current belief state is

$$\kappa \diamond dip = \{\{Blue\}, \{Red, Acid\}\}.$$

Since we also represent observations by sets of interpretations, the observation that the paper is red is given by the set $\alpha$ defined as follows

$$\alpha = \{\{Red, Acid\}, \{Red\}, \{Red, Blue, Acid\}, \{Red, Blue\}\}.$$

Note that this observation is consistent with the current belief state, because the intersection is non-empty. Therefore, by [AGM3] and [AGM4], it follows that any AGM revision operator will define the revised belief state to be this intersection. Hence, for any AGM revision operator $*$, the final belief state is

$$\{\{Red, Acid\}\}.$$

So, after dipping the litmus paper and observing the paper is red, we correctly believe that the beaker contains an acid.

## 3. Belief Update Preceding Belief Revision

As stated previously, we are interested in the belief change due to an alternating sequence of actions and observations. The easiest example consists of a single action followed by a single observation. However, throughout this paper, we assume that actions are infallible, and that actions are infallible and the effects of actions are completely specified. Under this assumption, a sequence of actions is no more difficult to handle than a single action. As such, the simplest interesting case to consider is given by an expression of the form

$$\kappa \diamond A_1 \diamond \cdots \diamond A_n * \alpha. \tag{2}$$

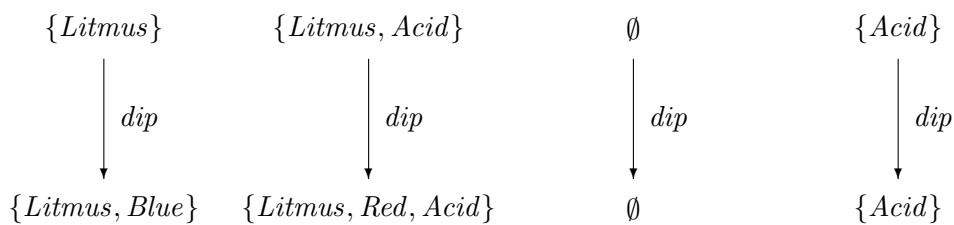

Figure 2: Extended Litmus Test

In this case, since there is only a single observation, we can focus entirely on the interaction between revision and update. The general case involving several observations is complicated by the fact that it implicitly requires some form of iterated revision. Our formalism handles the general case, but our initial focus will be on problems of the form in (2). In the next section, we consider an example of such a problem and we illustrate that the most natural solution requires the initial state to be revised at a later point in time.

We use the phrase *iterated belief change* to refer to any scenario where the beliefs of an agent change as a result of multiple sequential events. This is a somewhat non-standard use of the term, as most of the literature on iterated belief change is concerned with the (difficult) problem of *iterated belief revision*. By contrast, we consider problems of the form 2 to be instances of iterated belief change because a sequence of events is leading to a change in beliefs.

### 3.1 The Extended Litmus Paper Problem

We extend the litmus paper problem. The extended problem is just like the original, except that we allow for the possibility that the paper is not litmus paper; it might simply be a piece of plain white paper. In order to provide a formal representation of this problem, we need to extend the transition system used to represent the original problem by introducing a new fluent symbol *Litmus*. Informally, *Litmus* is true just in case the paper is actually litmus paper.

The action signature $\langle \mathbf{A}, \mathbf{F} \rangle$ consists of $\mathbf{A} = \{dip\}$ and $\mathbf{F} = \{Red, Blue, Acid, Litmus\}$. We assume the transition system for action effects given in Figure 2, and we assume a given AGM revision operator $*$.

We describe a sequence of events informally. Initially, the agent believes that the paper is a piece of litmus paper, but the agent is unsure about the contents of the beaker. To test the contents, the agent dips the paper in the beaker. After dipping, the agent looks at the paper and observes that it is still white. We are interested in determining a plausible final belief state.

We now give a more formal representation of the problem. The initial belief state is

$$\kappa = \{\{Litmus\}, \{Litmus, Acid\}\}.$$

After dipping the paper in the beaker, we update the belief state as follows:

$$\kappa \diamond dip = \{\{Litmus, Blue\}, \{Litmus, Red, Acid\}\}.$$

At this point, the agent looks at the paper and sees that it is neither blue nor red. This observation is represented by the following set of worlds:

$$\alpha = \{\emptyset, \{Litmus\}, \{Acid\}, \{Litmus, Acid\}\}.$$

The naive suggestion is to simply revise $\kappa \diamond dip$ by $\alpha$. However, this is not guaranteed to give the correct result. It is possible, for example, to define an AGM operator which gives the following final belief state:

$$\kappa' = \{\{Litmus\}, \{Litmus, Acid\}\}.$$

In this case, the agent believes that the piece of paper is white litmus paper. This is clearly not a plausible final belief state.

Informally, if the paper is litmus paper, then it must be either red or blue after a dipping action is performed. Hence, neither $\{Litmus\}$ nor $\{Litmus, Acid\}$ is a plausible state after dipping; simply revising by the observation may give a belief state that is incorrect because the revision operator does not encode this constraint. The final belief state should consist entirely of states that are possible consequences of dipping. Even if some particular AGM operator does give a plausible final belief state for this example, the process through which the final beliefs is obtained is not sound. In this kind of example, an observation after dipping is actually giving information about the initial belief state. As such, our intuition is that a rational agent should revise the *initial belief state* in any case.

We suggest that a rational agent should reason as follows. After dipping the paper and seeing that it does not change colour, an agent should conclude that the paper was never litmus paper to begin with. The initial belief state should be modified to reflect this new belief *before* calculating the effects of the dipping action. This approach ensures that we will have a final belief state that is a possible outcome of dipping. At the end of the experiment, the agent should believe that the paper is not litmus paper and the agent should have no definite beliefs regarding the contents of the beaker. Hence, we propose that the most plausible final belief state is the set

$$\{\emptyset, \{Acid\}\}.$$

This simple example serves to illustrate the fact that it is sometimes useful for an agent to revise prior belief states in the face of new knowledge. In order to formalize this intuition in greater generality, we need to introduce some new formal machinery.

### 3.2 Interaction Between Revision and Update

In this section, we give a set of formal properties that we expect to hold when an update is followed by a revision. The properties are not overly restrictive and they do not provide a basis for a categorical semantics; they simply provide a point for discussion and comparison. Our underlying assumption is that action histories are infallible. The most recent observation is always incorporated, provided that it is consistent with the history of actions that have been executed. Hence, the properties we discuss are only expected to hold in action domains in which there are no failed actions and no exogenous actions.

We briefly present some of our underlying intuitions. Let $\kappa$ be a belief state, let $\bar{A}$ be a sequence of actions, and let $\alpha$ be an observation. We are interested in the situation where an agent has initial belief state $\kappa$, then $\bar{A}$ is executed, and then it is observed that the actual state must be in $\alpha$. We adopt the shorthand notation $\kappa \diamond \bar{A}$ as an abbreviation for the sequential update of $\kappa$ by each element of $\bar{A}$. There are three distinct cases to consider.

1. There are some $\alpha$-states in $\kappa \diamond \bar{A}$.

2. There are no $\alpha$-states in $\kappa \diamond \bar{A}$, but some $\alpha$-states in $2^{\mathbf{F}} \diamond \bar{A}$.

3. No $\alpha$-states are possible after executing $\bar{A}$.

Case (1) is the situation in which the observation $\alpha$ allows the agent to refine their knowledge of the world. After the observation $\alpha$, the agent should believe that the most plausible states are the states in $\kappa \diamond \bar{A}$ that are also in $\alpha$. In other words, the agent should adopt the belief state $(\kappa \diamond \bar{A}) \cap \alpha$.

In case (2), the agent should conclude that the actual state was not initially in $\kappa$. This conclusion is based on our underlying assumption that the action sequence $\bar{A}$ cannot fail, and the additional assumption that a new observation should be incorporated whenever possible. Both of these assumptions can be satisfied by modifying the initial belief state before performing the update. Informally, we would like to modify the initial belief state minimally in a manner that ensures that $\alpha$ will be true after executing $\bar{A}$. This is the case that occurs in the extended litmus paper problem.

Case (3) is problematic, because it suggests that the agent has some incorrect information: either the observation $\alpha$ is incorrect or the sequence $\bar{A}$ is incorrect. We are assuming that action histories are infallible. As such, the observation $\alpha$ must weakened in some manner in order to remain consistent with $\bar{A}$. In cases where there is no natural weakening, it may be necessary to abandon the observation completely. For a single observation, this is the approach that we take. We view a single observation as a disjunctive constraint on the possible states of world, and we assume that an observation has no meaning with respect to the non-member states. As such, if an agent discovers that the actual state of the world is not included in an observation, then that observation does not offer any information at all. When we consider multiple observations, we taken a more flexible approach which allows an agent to select a minimal repair of the sequence of observations.

Let $\kappa$ and $\alpha$ be sets of states, let $\bar{A}$ be a sequence of actions, let $\diamond$ be an update operator as in Definition 2, and let $*$ be an AGM revision operator. We formalize our intuitions with the following conditions that should hold when an update is followed by a revision.

**Interaction Properties**

P1. If $(2^F \diamond \bar{A}) \cap \alpha \neq \emptyset$, then $\kappa \diamond \bar{A} * \alpha \subseteq \alpha$

P2. If $(2^F \diamond \bar{A}) \cap \alpha = \emptyset$, then $\kappa \diamond \bar{A} * \alpha = \kappa \diamond \bar{A}$

P3. $(\kappa \diamond \bar{A}) \cap \alpha \subseteq \kappa \diamond \bar{A} * \alpha$

P4. If $(\kappa \diamond \bar{A}) \cap \alpha \neq \emptyset$, then $\kappa \diamond \bar{A} * \alpha \subseteq (\kappa \diamond \bar{A}) \cap \alpha$

P5. $\kappa \diamond \bar{A} * \alpha \subseteq 2^F \diamond \bar{A}$

We give some motivation for each property. P1 is a straightforward AGM-type assertion that $\alpha$ must hold after revising by $\alpha$, provided $\alpha$ is possible after executing $\bar{A}$. P2 handles the situation where it is impossible to be in an $\alpha$-world after executing $\bar{A}$. In this case, we simply discard the observation $\alpha$. Together, P1 and P2 formalize the underlying assumption that there are no failed actions.

P3 and P4 assert that revising by $\alpha$ is equivalent to taking the intersection with $\alpha$, provided the intersection is non-empty. These are similar to the AGM postulates asserting that revisions correspond to expansions, provided the observation is consistent with the knowledge base.

P5 provides the justification for revising prior belief states in the face of new knowledge. It asserts that, after revising by $\alpha$, we must still have a belief state that is a possible consequence of executing $\bar{A}$. In some cases, the only way to ensure that $\alpha$ holds after executing $\bar{A}$ is to modify the initial belief state. We remark that P5 does not indicate how the initial belief state should be modified.

It is worth noting that the interaction properties make no mention of *minimal change* with respect to the belief state $\kappa$. There is a notion of minimal change that is implicit in the revision operator $*$, but this is a notion change with respect to a measure plausibility that is completely independent of the stated properties.

## 3.3 Representing Histories

Transition systems are only suitable for representing *Markovian* action effects; that is, action effects that depend only on the action executed and the current state of the world. However, in the extended litmus paper problem, we saw that the outcome of an observation may depend on prior belief states. Even if action effects are Markovian, it does not follow that changes in belief are Markovian. As such, we need to introduce some formal machinery for representing histories. We will be interested in the historical evolution of an agent's beliefs, along with all of the actions executed. In order to do so, we need to introduce *trajectories* of belief states, observations, and actions. For a given action signature $\langle A, F \rangle$, we will use the following terminology.

1. A *belief trajectory* is an $n$-tuple $\langle \kappa_0, \ldots, \kappa_{n-1} \rangle$ of belief states.

2. An *observation trajectory* is an $n$-tuple $\bar{\alpha} = \langle \alpha_1, \ldots, \alpha_n \rangle$ where each $\alpha_i \in 2^S$.

3. An *action trajectory* is an $n$-tuple $\bar{A} = \langle A_1, \ldots, A_n \rangle$ where each $A_i \in \mathbf{A}$.

Note that, as a matter of convention, we start the indices at 0 for belief trajectories and we start the indices at 1 for observation and action trajectories. The rationale for this convention will be clear later. We also adopt the convention hinted at in the definitions, whereby the $i^{th}$ component of an observation trajectory $\bar{\alpha}$ will be denoted by $\alpha_i$, and the $i^{th}$ component of an action trajectory $\bar{A}$ will be denoted by $A_i$.

We remark that a belief trajectory is an agent's subjective view of how the world has changed. Hence, a belief trajectory represents the agent's current beliefs about the world history, not a historical account of what an agent believed at each point in time. For example, in the extended litmus paper problem, at the end of the experiment the agent believes that they were never holding a piece of litmus paper. The fact that the agent once believed that they were holding litmus paper is a different issue, one that is not represented in our formal conception of a belief trajectory.

We define a notion of consistency between observation trajectories and action trajectories. The intuition is that an observation trajectory $\bar{\alpha}$ is consistent with an action trajectory $\bar{A}$ if and only if each observation $\alpha_i$ is possible, given that the actions $(A_j)_{j \leq i}$ have been executed.

**Definition 3** *Let* $\bar{\alpha} = \langle \alpha_1, \ldots, \alpha_n \rangle$ *be an observation trajectory and let* $\bar{A} = \langle A_1, \ldots, A_n \rangle$ *be an action trajectory. We say that* $\bar{A}$ *is consistent with* $\bar{\alpha}$ *if and only if there is a belief trajectory* $\langle \kappa_0, \ldots, \kappa_n \rangle$ *such that, for all* $i$ *with* $1 \leq i \leq n$,

   *1.* $\kappa_i \subseteq \alpha_i$

   *2.* $\kappa_i = \kappa_{i-1} \diamond A_i$.

If $\bar{A}$ is consistent with $\bar{\alpha}$, we write $\bar{A} || \bar{\alpha}$.

A pair consisting of an action trajectory and an observation trajectory gives a complete picture of an agent's view of the history of the world. As such, it is useful to introduce the following terminology.

**Definition 4** *A world view of length* $n$ *is a pair* $W = \langle \bar{A}, \bar{\alpha} \rangle$, *where* $\bar{A}$ *is an action trajectory and* $\bar{\alpha}$ *is an observation trajectory, each of length* $n$. *We say* $W$ *is consistent if* $\bar{A} || \bar{\alpha}$.

## 4. Belief Evolution

We are interested in providing a formal treatment of alternating action/observation sequences of the form

$$\kappa \diamond A_1 * \alpha_1 \diamond \cdots \diamond A_n * \alpha_n. \tag{3}$$

Note that there is an implicit tendency to associate the operators in an expression of this form from left to right, which gives the following expression:

$$(\ldots ((\kappa \diamond A_1) * \alpha_1) \diamond \cdots \diamond A_n) * \alpha_n. \tag{4}$$

The extended litmus paper problem illustrates that this association can lead to unsatisfactory results. As such, we would like to propose an alternative method for evaluating expressions of the form (3). However, discussing expressions of this form directly can be somewhat misleading, since we can not preclude the interpretation in (4). In order to make it explicit that we are not computing successive updates and revisions, we introduce a new *belief evolution* operator $\circ$. Intuitively, $\circ$ simply allows us to associate the information to be incorporated in a manner closer to the following informal expression:

$$\kappa \circ (A_1, \alpha_1, \ldots, A_n, \alpha_n). \tag{5}$$

Note that the update and revision operators have disappeared; we are left with an expression that groups all actions and observations as a single sequence, suggesting that all information should be incorporated simultaneously. However, the order of observations and actions is still significant. Moreover, it is important to keep in mind that ∘ is defined with respect to given update and revision operators. The new operator is introduced primarily to give us a formal tool to make it explicit that we are not interpreting expressions of the form (3) by the default interpretation in (4).

The formal definition of ∘ is presented in the following sections, and it does not use the exact syntax in the informal expression (5). In the actual definition, a belief evolution operator takes a belief state and a world view as arguments. Also, the value returned is not a single belief state; it is a belief trajectory. However, (5) provides the important underlying intuition.

### 4.1 Infallible Observations

In this section, we define ∘ under the assumption that observations are always correct. Formally, this amounts to a restriction on the world views that are considered. In particular, we need not consider inconsistent world views. It is easy to see that an inconsistent world view is not possible under the assumption that action histories and observations are both infallible.

Let $s^{-1}(A)$ denote the set of all states $s'$ such that $(s', A, s) \in R$. We call $s^{-1}(A)$ the *pre-image* of $s$ with respect to $A$. The following definition generalizes this idea to give the pre-image of a set of states with respect to a sequence of actions.

**Definition 5** *Let $T$ be a deterministic transition system, let $\bar{A} = \langle A_1, \ldots, A_n \rangle$ and let $\alpha$ be an observation. Define $\alpha^{-1}(\bar{A}) = \{s \mid s \leadsto_{\bar{A}} s' \text{ for some } s' \in \alpha\}$.*

Hence, if the actual world is an element of $\alpha$ following the action sequence $\bar{A}$, then the initial state of the world must be in $\alpha^{-1}(\bar{A})$.

For illustrative purposes, it is useful to consider world views of length 1. Suppose we have an initial belief state $\kappa$, an action $A$ and an observation $\alpha$. Without formally defining the belief evolution operator ∘, we can give an intuitive interpretation of an expression of the form

$$\kappa \circ \langle \langle A \rangle, \langle \alpha \rangle \rangle = \langle \kappa_0, \kappa_1 \rangle.$$

The agent knows that the actual world is in $\alpha$ at the final point in time, so we must have $\kappa_1 \subseteq \alpha$. Moreover, the agent should believe that $\kappa_1$ is a possible result of executing $A$ from $\kappa_0$. In other words, we must have $\kappa_0 \subseteq \alpha^{-1}(A)$. All other things being equal, the agent would like to keep as much of $\kappa$ as possible. In order to incorporate $\alpha^{-1}(A)$ while keeping as much of $\kappa$ as possible, the agent should revise $\kappa$ by $\alpha^{-1}(A)$. This suggests the following solution.

1. $\kappa_0 = \kappa * \alpha^{-1}(A)$,

2. $\kappa_1 = \kappa_0 \diamond A$.

This reasoning can be applied to world views of length greater than 1. The idea is to trace every observation back to a precondition on the initial belief state. After revising the initial belief state by all preconditions, each subsequent belief state can be determined by a standard update operation.

We have the following formal definition for $\circ$. In the definition, for $i \leq n$ we let $\bar{A}_i$ denote the subsequence of actions $\langle A_1, \ldots, A_i \rangle$.

**Definition 6** *Let $\kappa$ be a belief state, let $\diamond$ be an update operator, let $*$ be an AGM revision operator, let $\bar{A}$ be an action trajectory of length $n$ and let $\bar{\alpha}$ be an observation trajectory of length $n$ such that $\bar{A}||\bar{\alpha}$. Define*

$$\kappa \circ \langle \bar{A}, \bar{\alpha} \rangle = \langle \kappa_0, \ldots, \kappa_n \rangle$$

*where*

1. *$\kappa_0 = \kappa * \bigcap_i \alpha_i^{-1}(\bar{A}_i)$*

2. *for $i \geq 1$, $\kappa_i = \kappa_0 \diamond A_1 \diamond \cdots \diamond A_i$.*

We remark that the intersection of observation preconditions in the definition of $\kappa_0$ is non-empty, because $\bar{A}||\bar{\alpha}$.

The following propositions are immediate, and they demonstrate that for some action sequences of length 1, the operator $\circ$ reduces to either revision or update. In each proposition, we assume that $\bar{A}||\bar{\alpha}$.

**Proposition 1** *Let $\kappa$ be a belief state, let $\bar{A} = \langle A \rangle$ and let $\bar{\alpha} = \langle 2^{\mathbf{F}} \rangle$. Then*

$$\kappa \circ \langle \bar{A}, \bar{\alpha} \rangle = \langle \kappa, \kappa \diamond A \rangle.$$

**Proof** Recall that we assume every action is executable in every state. It follows that $(2^{\mathbf{F}})^{-1}(A) = 2^{\mathbf{F}}$. Therefore

$$\begin{aligned} \kappa \circ \langle \bar{A}, \bar{\alpha} \rangle &= \langle \kappa * 2^{\mathbf{F}}, (\kappa * 2^{\mathbf{F}}) \diamond A \rangle \\ &= \langle \kappa, \kappa \diamond A \rangle. \end{aligned}$$

$\square$

In the next result, recall that $\lambda$ is a null action that never changes the state of the world.

**Proposition 2** *Let $\kappa$ be a belief state, let $\bar{A} = \langle \lambda \rangle$ and let $\bar{\alpha} = \langle \alpha \rangle$. Then*

$$\kappa \circ \langle \bar{A}, \bar{\alpha} \rangle = \langle \kappa * \alpha, \kappa * \alpha \rangle.$$

**Proof** Since $\lambda$ does not change the state, it follows that $\alpha^{-1}(\lambda) = \alpha$. Therefore

$$\begin{aligned} \kappa \circ \langle \bar{A}, \bar{\alpha} \rangle &= \langle \kappa * \alpha, (\kappa * \alpha) \diamond \lambda \rangle \\ &= \langle \kappa * \alpha, \kappa * \alpha \rangle. \end{aligned}$$

$\square$

Hence, the original revision and update operators can be retrieved from the $\diamond$ operator. As such, it is reasonable to compute the iterated belief change due to action in terms of belief evolution.

We will demonstrate that belief evolution provides a reasonable approach for computing the outcome of a sequence of actions and observations. We stress that computing updates and revisions in succession does not provide a reasonable solution in many cases. As such, we want to *define* the outcome of a sequence of updates and revisions in terms of a belief evolution operator. Given $*$ and $\diamond$, we *define* the iterated belief change

$$\kappa \diamond A * \alpha$$

to be the final belief state in the belief trajectory

$$\kappa \circ \langle \langle A \rangle, \langle \alpha \rangle \rangle.$$

We adopt this somewhat confusing convention temporarily in order to prove that belief evolution provides a semantics for iterated belief change that satisfies our interaction properties.

More generally, consider a sequence $\bar{A}$ of $n$ actions followed by a single observation $\alpha$. In this case, define $\bar{\alpha}_n$ to be the observation trajectory consisting of $n-1$ instances of $2^{\mathbf{F}}$ followed by the final observation $\alpha$. We define the iterated belief change

$$\kappa \diamond \bar{A} * \alpha$$

to be the final belief state in the belief trajectory

$$\kappa \circ \langle \bar{A}, \bar{\alpha}_n \rangle.$$

**Proposition 3** *Let $\bar{A}$ be an action trajectory and let $\alpha$ be an observation. If $\bar{A} || \bar{\alpha}_n$, the iterated belief change $\kappa \diamond \bar{A} * \alpha$ defined as above satisfies the interaction properties P1-P5.*

**Proof** Let $\kappa$ be a belief state. By the convention outlined above,

$$\kappa \diamond \bar{A} * \alpha = (\kappa * \alpha^{-1}(\bar{A})) \diamond \bar{A}.$$

We demonstrate that this definition satisfies P1-P5.

P1. If $(2^F \diamond \bar{A}) \cap \alpha \neq \emptyset$, then $(\kappa \diamond \bar{A}) * \alpha \subseteq \alpha$.
Note that the antecedent is true because $\bar{A}$ and $\bar{\alpha}_n$ are consistent. We have the following inclusions:

$$(\kappa * \alpha^{-1}(\bar{A})) \diamond \bar{A} \ \subseteq \ \alpha^{-1}(\bar{A}) \diamond \bar{A} \ \subseteq \ \alpha.$$

The first inclusion holds by [AGM2] plus the fact that update satisfies $(X \cap Y) \diamond \bar{A} \subseteq X \diamond \bar{A}$. The second inclusion holds by definition of the pre-image. Hence, the consequent is true.

P2. If $(2^F \diamond \bar{A}) \cap \alpha = \emptyset$, then $(\kappa \diamond \bar{A}) * \alpha = \kappa \diamond \bar{A}$
The antecedent is false, since $\bar{A}$ and $\bar{\alpha}_n$ are consistent.

**P3.** $(\kappa \diamond \bar{A}) \cap \alpha \subseteq (\kappa \diamond \bar{A}) * \alpha$

Suppose $s \in (\kappa \diamond \bar{A}) \cap \alpha$. So $s \in \alpha$ and there is some $s' \in \kappa$ such that $\bar{A}$ maps $s'$ to $s$. Hence, $s' \in \alpha^{-1}(\bar{A})$. It follows from [AGM2] that $s' \in \kappa * \alpha^{-1}(\bar{A})$. Since $\bar{A}$ maps $s'$ to $s$, we have $s \in (\kappa * \alpha^{-1}(\bar{A})) \diamond \bar{A}$.

**P4.** If $(\kappa \diamond \bar{A}) \cap \alpha \neq \emptyset$, then $(\kappa \diamond \bar{A}) * \alpha \subseteq (\kappa \diamond \bar{A}) \cap \alpha$

Suppose that $(\kappa \diamond \bar{A}) \cap \alpha \neq \emptyset$. So there is a state in $\kappa$ that is mapped to $\alpha$ by the sequence $\bar{A}$. Hence $\kappa \cap \alpha^{-1}(\bar{A}) \neq \emptyset$. By [AGM3] and [AGM4], it follows that $\kappa * \alpha^{-1}(\bar{A}) = \kappa \cap \alpha^{-1}(\bar{A})$. Now suppose that $s \in (\kappa * \alpha^{-1}(\bar{A})) \diamond \bar{A}$. So there exists $s' \in \kappa * \alpha^{-1}(\bar{A})$ such that $\bar{A}$ maps $s'$ to $s$. But then $s' \in \kappa \cap \alpha^{-1}(\bar{A})$. But this implies that $s \in \kappa \diamond \bar{A}$ and $s \in \alpha$.

**P5.** $(\kappa \diamond \bar{A}) * \alpha \subseteq 2^F \diamond \bar{A}$

This is immediate, because $(\kappa * \alpha^{-1}(\bar{A})) \subseteq 2^F$ and $(X \cup Y) \diamond \alpha = (X \diamond \alpha) \cup (Y \diamond \alpha)$. $\square$

The three preceding propositions demonstrate the suitability of $\circ$ as a natural operator for reasoning about the interaction between revision and update.

We can use a belief evolution operator to give an appropriate treatment of the litmus paper problem.

**Example**  Consider the extended litmus paper problem, and let

$$\alpha = \{\emptyset, \{Litmus\}, \{Acid\}, \{Litmus, Acid\}\}.$$

Hence, the world view $W = \langle \langle dip \rangle, \langle \alpha \rangle \rangle$ represents a dipping action followed by the observation that the paper is still white. If $\circ$ is obtained from the metric transition system defined by the Hamming distance and the transitions in Figure 2, the final belief state in $\kappa \circ W$ is given by

$$
\begin{aligned}
\kappa * \alpha^{-1}(dip) \diamond dip &= \kappa * \{\emptyset, \{Acid\}\} \diamond dip \\
&= \{\emptyset, \{Acid\}\}.
\end{aligned}
$$

This calculation is consistent with our original intuitions, in that the agent revises the initial belief state before updating by the *dip* action. This ensures that we will have a final belief state that is a possible outcome of dipping. Moreover, the initial belief state is revised by the pre-image of the final observation, which means it is modified as little as possible while still guaranteeing that the final observation will be feasible. Note also that the final belief state given by this calculation is intuitively reasonable. It simply indicates that the contents of the beaker are still unknown, but the agent now believes the paper is not litmus paper. Hence, a belief evolution operator employs a plausible procedure and returns a desirable result.

## 4.2 Fallible Observations

In this section, we consider belief evolution in the case where observations may be incorrect. Formally, this means that we are interested in determining the outcome of belief evolution for inconsistent world views. In this case, we cannot simply take the intersection of all observation pre-images, because this intersection may be empty. The basic idea behind our approach is to define belief evolution with respect to some external notion of the *reliability* of an observation.

We start by defining belief evolution in the most general case, with respect to a total pre-order over the elements of an observation trajectory $\bar{\alpha}$. In order to define the pre-order, we assume we are given a function $r$ that maps each element of $\bar{\alpha}$ to an integer. The actual value of $r(\alpha_i)$ is not particularly important; the function $r$ is just used to impose an ordering on the observations. We interpret

$$r(\alpha_i) < r(\alpha_j)$$

to mean that $\alpha_i$ is more reliable than $\alpha_j$. In this case, if consistency can be restored by discarding either $\alpha_i$ or $\alpha_j$, then $\alpha_j$ is discarded. The ranking function on observations may be obtained in several ways. For example, it may be induced from an ordering over all possible observations, indicating the reliability of sensing information. On the other hand, this pre-order might simply encode a general convention for dealing with sequential observations. For example, in some cases it may be reasonable to prefer a more recent observation over an earlier observation.

Our basic approach is the following. Given an observation trajectory $\bar{\alpha}$ that is not consistent with $\bar{A}$, we discard observations in a manner that gives us a consistent world view. To be precise, "discarding" an observation in this context means that we replace it with $2^{\mathbf{F}}$. We discard observations rather than weaken them, because we view the content of an observation as an atomic proposition. We are guided by two basic principles. First, all other things being equal, we would like to keep as much of $\bar{\alpha}$ as consistently possible. Second, when observations must be discarded, we try to keep those that are the most reliable. With this informal sketch in mind, we define $w(\bar{\alpha})$, the set of all trajectories obtained by discarding some of the observations in $\bar{\alpha}$.

**Definition 7** *Let $\bar{\alpha}$ be an observation trajectory of length $n$, and let $\mathcal{O}_n$ be the set of all observation trajectories of length $n$. Then define:*

$$w(\bar{\alpha}) = \{\bar{\alpha}' \mid \bar{\alpha}' \in \mathcal{O}_n \text{ and for } 1 \leq i \leq n, \alpha_i' = \alpha_i \text{ or } \alpha_i' = 2^{\mathbf{F}}\}.$$

We are interested in finding those trajectories in $w(\bar{\alpha})$ that are consistent with $\bar{A}$, while differing minimally from $\bar{\alpha}$. In the following definition, for observation trajectories of equal length $n$, we write $\bar{\alpha} \subseteq \bar{\alpha}'$ as a shorthand notation to indicate that $\alpha_i \subseteq \alpha_i'$ for every $1 \leq i \leq n$.

**Definition 8** *For a world view $\langle \bar{A}, \bar{\alpha} \rangle$:*

$$\langle \bar{A}, \bar{\alpha} \rangle \downarrow_\perp = \{\bar{\alpha}' \in w(\bar{\alpha}) \mid \bar{A}||\bar{\alpha}' \text{ and for all } \bar{\alpha}'' \in w(\bar{\alpha}) \text{ such that } \bar{\alpha}'' \subseteq \bar{\alpha}', \text{ not } \bar{A}||\bar{\alpha}''\}.$$

So $\bar{\alpha}' \in \langle \bar{A}, \bar{\alpha} \rangle \downarrow_\perp$ if and only if $\bar{\alpha}'$ is consistent with $\bar{A}$, but it becomes inconsistent if any of the discarded observations are re-introduced. Therefore the elements of $\bar{\alpha}' \in \langle \bar{A}, \bar{\alpha} \rangle \downarrow_\perp$ differ minimally from $\bar{\alpha}$, where "minimal" is defined in terms of *set-containment*.[1]

We can use the reliability ordering over observations to define a reliability ordering over observation trajectories.

---

1. Note that this is not the only reasonable notion of minimality that could be employed here. One natural alternative would be to consider minimal change in terms of *cardinality*. In this case, a trajectory differs

**Definition 9** *For $\bar{\alpha}', \bar{\alpha}'' \in \langle \bar{A}, \bar{\alpha} \rangle \downarrow_\perp$, we write $\bar{\alpha}'' < \bar{\alpha}'$ if and only if there is a $j$ such that*

1. *For all $k < j$ and all $i < n$, if $r(\alpha_i) = k$ then $\alpha_i' = \alpha_i''$.*

2. *There exists some $\alpha_i$ such that $r(\alpha_i) = j$ and $\alpha_i'' \subset \alpha_i'$.*

3. *There does not exist any $\alpha_i$ such that $r(\alpha_i) = j$ and $\alpha_i' \subset \alpha_i''$.*

Informally, $\bar{\alpha}'' < \bar{\alpha}'$ if $\bar{\alpha}''$ retains more reliable observations than $\bar{\alpha}'$. The minimal trajectories in this ordering are, therefore, those that retain the most plausible observations.

**Definition 10** *The set of repairs of a world view $\langle \bar{A}, \bar{\alpha} \rangle$ with respect to a reliability function $r$ is given by:*

$$Rep(\bar{A}, \bar{\alpha}) = \{ \bar{\alpha}' \in \langle \bar{A}, \bar{\alpha} \rangle \downarrow_\perp \mid \text{ there is no } \bar{\alpha}'' \in \langle \bar{A}, \bar{\alpha} \rangle \downarrow_\perp \text{ such that } \bar{\alpha}'' < \bar{\alpha}' \}.$$

Note that $Rep(\bar{A}, \bar{\alpha})$ may contain several observation trajectories. Moreover, each trajectory in $Rep(\bar{A}, \bar{\alpha})$ is consistent with $\bar{A}$, while minimally discarding observations and keeping the most reliable observations possible.

In Definition 6, we defined $\circ$ for consistent world views. For inconsistent world views, we use the following definition.

**Definition 11** *Let $\kappa$ be a belief state, let $\bar{A}$ be an action history, let $\bar{\alpha}$ be an observation trajectory, and let $r$ be a reliability function. If $\bar{A}$ is not consistent with $\bar{\alpha}$, then:*

$$\kappa \circ \langle \bar{A}, \bar{\alpha} \rangle = \{ \kappa \circ \langle \bar{A}, \bar{\alpha}' \rangle \mid \bar{\alpha}' \in Rep(\bar{A}, \bar{\alpha}) \}.$$

This definition is well-formed, because $\bar{\alpha}' \in Rep(\bar{A}, \bar{\alpha})$ implies $\bar{A} || \bar{\alpha}'$. Note that the outcome of belief evolution in this case is a set of belief trajectories.

We adopt the following convention. If $\langle \bar{A}, \bar{\alpha} \rangle = \{ \bar{\alpha}' \}$, then we write $\langle \bar{A}, \bar{\alpha} \rangle = \bar{\alpha}'$. If $|Rep(\bar{A}, \bar{\alpha})| = 1$, then belief evolution yields a unique belief trajectory. There are natural examples when this is the case.

**Proposition 4** *Let $\langle \bar{A}, \bar{\alpha} \rangle$ be a world view of length $n$. Let $r$ be a reliability function such that $\alpha_i \neq \alpha_j$ implies $r(\alpha_i) \neq r(\alpha_j)$ for all $1 \leq i, j \leq n$. Then $|Rep(\bar{A}, \bar{\alpha})| = 1$.*

**Proof** Note that $\langle \bar{A}, \bar{\alpha} \rangle \downarrow_\perp \neq \emptyset$, because it is always possible to find a trajectory consistent with $\bar{A}$ by discarding some of the observations in $\bar{\alpha}$. It follows immediately that $Rep(\bar{A}, \bar{\alpha})$ is non-empty. Hence $|Rep(\bar{A}, \bar{\alpha})| \geq 1$.

Now suppose that there exist $\bar{\alpha}'$ and $\bar{\alpha}''$ such that $\bar{\alpha}' \in Rep(\bar{A}, \bar{\alpha})$, $\bar{\alpha}'' \in Rep(\bar{A}, \bar{\alpha})$ and $\bar{\alpha}' \neq \bar{\alpha}''$. Thus

$$\{ \alpha_i \mid \alpha_i' \neq \alpha_i'' \} \neq \emptyset.$$

Let $\alpha_j \in \{ \alpha_i \mid \alpha_i' \neq \alpha_i'' \}$ be such that $r(\alpha_j)$ is minimal. By assumption, $\alpha_j$ is unique. If $\alpha_j' = \alpha_j$, then $\bar{\alpha}' < \bar{\alpha}''$ which contradicts $\bar{\alpha}'' \in Rep(\bar{A}, \bar{\alpha})$. If $\alpha_j' = 2^{\mathbf{F}}$, then $\bar{\alpha}'' < \bar{\alpha}'$ which

---

minimally from $\bar{\alpha}$ just in case a minimal number of observations is discarded. There are reasonable arguments for both the containment approach and the cardinality approach, depending on the context. Neither approach has a clear theoretical advantage, and the development of each is virtually identical. We determined that this paper is better served by presenting the containment approach in detail, rather than presenting a series of duplicate results for different conceptions of minimality.

contradicts $\bar\alpha' \in Rep(\bar{A}, \bar\alpha)$. Therefore, $\bar\alpha' = \bar\alpha''$. It follows that $|Rep(\bar{A}, \bar\alpha)| \leq 1$, because no two elements of $Rep(\bar{A}, \bar\alpha)$ can be distinct.  □

Thus if $r$ assigns a unique value to each observation, then belief evolution yields a unique outcome. We return to this fact in the next section.

In cases where belief evolution does not yield a unique result, a skeptical approach can be defined by taking a union on initial belief states. Recall that $\kappa_0$ is the first element in the trajectory $\bar\kappa$. If we define

$$\kappa_0 = \bigcup \{\kappa_0' \mid \overline{\kappa'} \in \kappa \circ Rep(\bar{A}, \bar\alpha)\}.$$

then a unique belief trajectory can then be defined by computing the effects of actions. This trajectory is general enough to include the outcome of every minimal repair. This kind of skeptical approach is appropriate in some situations.

## 4.3 Recency

One well-known approach to dealing with sequences of observations is to give precedence to *recent* information (Nayak, 1994; Papini, 2001). Given an observation trajectory $\bar\alpha$, a preference for recent information can be represented in our framework by defining $r$ such that $r(\alpha_i) = -i$. For our purposes, recency provides a concrete reliability ordering over observations, which facilitates the presentation of examples and comparison with related formalisms. As such, throughout the remainder of this paper, we use $\circ$ to denote the belief evolution operator $\circ$ defined with respect to this function $r$.

We stress that this preference for recent information is just a convention that we adopt because it simplifies the exposition, and we note that this convention has been the subject of criticism (Delgrande, Dubois, & Lang, 2006). Note that, by Proposition 4, belief evolution under the recency convention defines a unique belief trajectory as an outcome. As such, belief evolution under the recency convention also defines a specific approach to iterated revision. A sequence of several observations interspersed with null actions is no longer computed by simply applying a single shot revision operator several times. We will explore the approach to iterated revision that is implicit in our belief evolution operators in §6.2.

We conclude this section with a useful result. Thus far, applying the $\circ$ operator requires tracing action preconditions back to the initial state for revision, then applying action effects to get a complete history. If we are only concerned with the final belief state, then there are many cases in which we do not need to go to so much effort. In the following proposition, it is helpful to think of $2^{\mathbf{F}}$ as a "null observation" that provides no new information.

**Proposition 5** *Let $\kappa$ be a belief state, let $\bar{A}$ be an action trajectory of length $n$ and let $\alpha$ be a belief state such that $\alpha \subseteq \kappa \diamond \bar{A}$. If $\bar\alpha$ is the observation trajectory with $n-1$ observations of $2^{\mathbf{F}}$ followed by a single observation $\alpha$, then the final belief state in $\kappa \circ \langle \bar{A}, \bar\alpha \rangle$ is $(\kappa \diamond \bar{A}) * \alpha$.*

**Proof**  By definition, the final belief state of $\kappa \circ \langle \bar{A}, \bar\alpha \rangle$ is

$$(\kappa * \alpha^{-1}(\bar{A})) \diamond \bar{A}.$$

Since $\alpha \subseteq \kappa \diamond \bar{A}$, the intersection $\kappa \cap \alpha^{-1}(\bar{A})$ is non-empty. By [AGM3] and [AGM4], it follows that

$$\kappa * \alpha^{-1}(\bar{A}) = \kappa \cap \alpha^{-1}(\bar{A})$$

and therefore

$$(\kappa * \alpha^{-1}(\bar{A})) \diamond \bar{A} = (\kappa \cap \alpha^{-1}(\bar{A})) \diamond \bar{A}.$$

Clearly, the right hand side of this equality is equal to $(\kappa \diamond \bar{A}) \cap \alpha$. Again, since $\alpha \subseteq \kappa \diamond \bar{A}$, it follows from [AGM3] and [AGM4] that this is $(\kappa \diamond \bar{A}) * \alpha$. $\quad\square$

The proposition indicates that, given a single observation that is consistent with the actions that have been executed, we can simply revise the outcome of the actions and we get the correct final belief state.

## 5. Defining Belief Evolution Through Orderings on Interpretations

In the next two sections, we provide a characterization of belief evolution operators terms of total pre-orders on interpretations. We restrict attention to the case involving one action followed by one observation. This result extends easily to the case involving $n$ actions followed by one observation, because action sequences of length $n$ simply define a new set of transitions over states. Since we prove our characterization for an arbitrary action signature, allowing $n$ actions prior to a single observation is no more difficult than allowing only one action. We present the case with a single action, as it simplifies the exposition by allowing us to avoid introducing sequences of null observations interspersed with the actions. We remark that our result does not extend directly to the case involving several observations, as we do not introduce an axiomatic account of the reliability of an observation.

First, we need to delineate a general class of belief change functions.

**Definition 12** *A combined belief change operator is a function*

$$\dot{\diamond} : 2^S \times \langle \mathbf{A}, 2^S \rangle \to 2^S.$$

Hence, a combined belief change operator takes a belief state and an ordered pair $\langle A, \alpha \rangle$ as input, and it returns a new belief state.

For a fixed update operator $\diamond$ and a fixed revision operator $*$, consider the following postulates.

**I1** If $(2^F \diamond A) \cap \alpha \neq \emptyset$, then $\kappa \dot{\diamond} \langle A, \alpha \rangle = \kappa * \alpha^{-1}(A) \diamond A$.

**I2** If $(2^F \diamond A) \cap \alpha = \emptyset$, then $\kappa \dot{\diamond} \langle A, \alpha \rangle = \kappa \diamond A$.

If we abuse our notation by letting $\kappa \diamond \langle A, \alpha \rangle$ denote the final belief state in the corresponding belief trajectory, then we get the following result. In the proposition, we we refer to a the belief evolution operator *defined by* an update operator and a revision operator. To be clear, this refers to the belief evolution operator obtained through Definitions 6 and 11.

**Proposition 6** *Let $\diamond$ be a belief update operator and let $*$ be a belief revision operator. Then $\dot{\diamond}$ is the belief evolution operator defined by $\diamond, *$ if and only if $\dot{\diamond}$ satisfies **I1** and **I2**.*

**Proof** Let $\dot{\diamond}$ be the belief evolution operator corresponding to $\diamond$ and $*$. If $(2^F \diamond A) \cap \alpha \neq \emptyset$, then $\kappa \dot{\diamond} \langle A, \alpha \rangle = \kappa * \alpha^{-1}(A) \diamond A$ by definition. Hence $\dot{\diamond}$ satisfies **I1**. Suppose, on the other hand, that $(2^F \diamond A) \cap \alpha = \emptyset$. In this case $\alpha^{-1}(A) = \emptyset$. Therefore, $\kappa \dot{\diamond} \langle A, \alpha \rangle = \kappa * 2^F \diamond A = \kappa \diamond A$. So $\dot{\diamond}$ satisfies **I2**.

To prove the converse, suppose that $\dot{\diamond}$ satisfies **I1** and **I2**. Let $\circ$ be the belief evolution operator defined by $\diamond$ and $*$. Suppose that $(2^F \diamond A) \cap \alpha \neq \emptyset$. If follows that

$$
\begin{aligned}
\kappa \circ \langle A, \alpha \rangle &= \kappa * \alpha^{-1} \diamond A && \text{(since } A \text{ and } \alpha \text{ are consistent)} \\
&= \kappa \dot{\diamond} \langle A, \alpha \rangle && \text{(by **I1**)}
\end{aligned}
$$

Now suppose that $(2^F \diamond A) \cap \alpha = \emptyset$.

$$
\begin{aligned}
\kappa \circ \langle A, \alpha \rangle &= \kappa * 2^F \diamond A && \text{(since } A \text{ and } \alpha \text{ are not consistent)} \\
&= \kappa \dot{\diamond} \langle A, \alpha \rangle && \text{(by **I2**)}
\end{aligned}
$$

This completes the proof. $\square$

Hence, the postulates **I1** and **I2** provide a complete syntactic description of belief evolution. This characterization will make it easier to state the representation result in the next section.

## 5.1 Translations on Orderings

For a fixed transition system $T$, we would like to provide a characterization of all combined belief change functions that represent belief evolution operators for $T$. Our characterization will be defined in terms of total pre-orderings over interpretations. First we need to introduce the basic characterization of AGM revision operators in terms of total pre-orders, due to Katsuno and Mendelzon (1992). Our presentation differs slightly from the original because we define revision operators on sets of states rather than formulas.

**Definition 13** *(Katsuno & Mendelzon, 1992) Given a belief state $\kappa$, a total pre-order $\leq_\kappa$ over interpretations is called a* faithful ranking *with respect to $\kappa$ just in case the following conditions hold:*

- *If $s_1, s_2 \in \kappa$, then $s_1 =_\kappa s_2$.*

- *If $s_1 \in \kappa$ and $s_2 \notin \kappa$, then $s_1 <_\kappa s_2$.*

Hence, a faithful ranking is simply a total pre-order where the minimal elements are the members of $\kappa$. In order to simplify the discussion, we introduce the following notation. If $\alpha$ is a set of states and $\leq$ is an ordering on a superset of $\alpha$:

$$\min(\alpha, \leq) = \{s \mid s \in \alpha \text{ and } s \text{ is } \leq\text{-minimal among interpretations in } \alpha)\}.$$

The following definition characterizes AGM revision operators in terms of pre-orders on interpretations.

**Proposition 7** *(Katsuno & Mendelzon, 1992) Given a belief state $\kappa$, a revision operator $*$ satisfies the AGM postulates if and only if there exists a faithful ranking $\leq_\kappa$ with respect to $\kappa$ such that for any set of interpretations $\alpha$:*

$$\kappa * \alpha = \min(\alpha, \leq_\kappa).$$

In the remainder of this section, we extend this result to characterize belief evolution operators.

Assume we are given a fixed transition system $T = \langle S, R \rangle$ defining an update operator $\diamond$. The following definition gives a natural progression operation on pre-orderings.

**Definition 14** *If $\leq_\kappa$ is a faithful ranking with respect to $\kappa$ and $A$ is an action, then define $\leq_{\kappa \diamond A}$ such that $s_1 \leq_{\kappa \diamond A} s_2$ if and only if*

- *There exist $t_1, t_2$ such that $(t_1, A, s_1) \in R$ and $(t_2, A, s_2) \in R$).*

- $t_1 \leq_\kappa t_2$.

Note that $\leq_{\kappa \diamond A}$ is not generally a total pre-order because it may be the case that some states are not possible outcomes of the action $A$. Hence $\leq_{\kappa \diamond A}$ is a partial pre-order. We can think of $\leq_{\kappa \diamond A}$ as a "shifted" ordering, with minimal elements $\kappa \diamond A$.

The following definition associates a combined revision operator with a faithful ranking.

**Definition 15** *Let $\leq_\kappa$ be a faithful ranking with respect to $\kappa$. The combined belief change operator associated with $\leq_\kappa$ is the following:*

$$\kappa \dot\diamond \langle A, \alpha \rangle = \begin{cases} \min(\alpha, \leq_{\kappa \diamond A}) & \text{if } (2^F \diamond A) \cap \alpha \neq \emptyset \\ \kappa \diamond A & \text{otherwise.} \end{cases}$$

Note that the operator $\dot\diamond$ takes an observation and an action as inputs, and it returns a new belief state. We will prove that the class of functions definable in this manner coincides exactly with the class of belief evolution operators.

We first prove that every faithful ranking defines a belief evolution operator.

**Proposition 8** *Let $\leq_\kappa$ be a faithful ranking with respect to $\kappa$ and let $\dot\diamond$ be the combined belief change operator defined by $\leq_\kappa$. Then $\dot\diamond$ satisfies **I1** and **I2**.*

**Proof** Let $A$ be an action, let $\alpha$ be an observation, and let $*$ be the AGM revision operator defined by $\leq_\kappa$ by Proposition 7. We will prove that $\dot\diamond$ satisfies **I1** and **I2** with respect to $*$.

Suppose that $(2^F \diamond A) \cap \alpha \neq \emptyset$, so

$$\kappa \dot\diamond \langle A, \alpha \rangle = \min(\alpha, \leq_{\kappa \diamond A}).$$

We remark that, by definition,

$$\alpha = \alpha^{-1}(A) \diamond A.$$

So:

$$\begin{aligned} \kappa \dot\diamond \langle A, \alpha \rangle &= \min(\alpha^{-1}(A) \diamond A, \leq_{\kappa \diamond A}) \\ &= \min(\alpha^{-1}(A), \leq_\kappa) \diamond A \end{aligned}$$

By definition of $*$:

$$\kappa * \alpha^{-1}(A) = \min(\alpha^{-1}(A), \leq_\kappa)$$

It follows that

$$\kappa \dot\diamond \langle A, \alpha \rangle = (\kappa * \alpha^{-1}(A)) \diamond A$$

which proves that **I1** holds.

If $(2^F \diamond A) \cap \alpha = \emptyset$, then by definition:

$$\kappa \dot\diamond \langle A, \alpha \rangle = \kappa \diamond A.$$

Hence **I2** is satisfied. □

We now prove the converse.

**Proposition 9** *Let $\dot{\diamond}$ be an operator satisfying* **I1** *and* **I2** *for some AGM revision function* $*$. *Given a belief state* $\kappa$, *there is a faithful ranking* $\leq_\kappa$ *such that* $\dot{\diamond}$ *is the combined belief change operator defined by* $\leq_\kappa$.

**Proof**  By Proposition 7, there is a a faithful ranking $\leq_\kappa$ with respect to $\kappa$ such that

$$\kappa * \alpha = \min(\alpha, \leq_\kappa)$$

for all $\alpha$. Fix a particular $\alpha$ and let $A$ be an action symbol. Suppose that $(2^F \diamond A) \cap \alpha \neq \emptyset$. So, by **I1**:

$$\kappa \dot{\diamond} \langle A, \alpha \rangle = \kappa * \alpha^{-1}(A) \diamond A.$$

By definition of $*$, this is equal to

$$\min(\alpha^{-1}(A), \leq_\kappa) \diamond A$$

This is equivalent to:

$$\min(\alpha^{-1}(A) \diamond A, \leq_{\kappa \diamond A})$$

Simplifying the first argument, we get:

$$\min(\alpha, \leq_{\kappa \diamond A})$$

which is what we wanted to show.

Now suppose that $\alpha$ $(2^F \diamond A) \cap \alpha = \emptyset$ then,

$$\kappa \dot{\diamond} \langle A, \alpha \rangle \;\; = \;\; \kappa \diamond A \qquad \text{(by **I2**)}$$

This completes the proof.  $\square$

Hence, the class of belief evolution operators can be characterized by simply shifting the total pre-order that defines the revision operator. This characterization is essentially a corollary of Katsuno and Mendelzon's representation result for AGM revision. However, this approach does represent a significant departure from the iterative approach to applying update and revision operators. What our result demonstrates is that the progression of beliefs through actions should be applied at the level of *the same pre-order* that is used for revision. In this manner, the relative likelihood of all states is shifted appropriately when an action is executed. This ensures that later revisions use the same ordering that would be used initially, which captures our intuition that the execution of actions does not change any a priori likelihood of initial states of the world.

## 6. Comparison with Related Work

Many action formalisms define action effects to be Markovian. This is the case, for example, in action languages like $\mathcal{A}$ (Gelfond & Lifschitz, 1998). When action formalisms of this kind are supplemented with sensing actions, there is a natural tendency to compute epistemic change by computing the effects of ontic actions and sensing actions in succession. This is the implicit approach to iterated belief change caused by actions in the epistemic extensions of $\mathcal{A}$ (Lobo et al., 2001; Son & Baral, 2001). We have seen that this strategy is not appropriate for litmus-type problems. However, this does not mean that such formalisms can not be used for reasoning about iterated belief change due to action. It simply means that some care must be taken to define the iterated change correctly.

Belief evolution should not be seen as a formalism in competition with Markovian formalisms; it should be seen as a methodology for extending Markovian formalisms to address iterated belief change. If the revision and update operators are given explicitly, the definition of the corresponding belief evolution operator is straightforward. This is true even in formalisms where the basic operators are relatively sophisticated, such as those defined in the multi-agent belief structures of Herzig, Lang and Marquis (2004).

## 6.1 The Situation Calculus

In this section, we compare our work with an extended version of the SitCalc that explicitly represents the beliefs of agents. We assume the reader is familiar with the SitCalc (Levesque, Pirri, & Reiter, 1998), and we provide only a brief introduction of the notation we will use.

The terms in a SitCalc action theory range over a domain that includes *situations* and *actions*. A *fluent* is a predicate that takes a situation as the final argument. A SitCalc action theory includes an *action precondition* axiom for each action symbol, a *successor state axiom* for each fluent symbol, as well as the foundational axioms of the SitCalc. Informally, a situation represents the state of the world, along with a complete history of actions that have been executed. There is a distinguished constant $S_0$ that represents the initial situation, and a distinguished function symbol *do* that represents the execution of an action. Every situation term can be written as follows:

$$do(A_n, do(A_{n-1}, \ldots, do(A_1, S_0) \ldots)).$$

To simplify the notation, we will abbreviate this situation as $do([A_1, \ldots, A_n], S_0)$.

To clarify the results that follow, we adopt the following convention. Expressions such as $s$ and $do(A, s)$ will be used as syntactic variables ranging over *situation terms*. We will also need to refer explicitly to the semantic objects denoted by terms in a given first-order interpretation $\mathcal{M}$ of a situation calculus theory. Hence, we let $s^{\mathcal{M}}$ denote the *situation* denoted by the *situation term* $s$ in the first-order interpretation $\mathcal{M}$. We adopt the same convention to denote the extensions of predicate symbols in an interpretation.

The SitCalc has been extended to include a representation of belief (Shapiro et al., 2000). The extension includes a distinguished fluent symbol $B$ that represents an *accessibility relation* on situations, similar to those used in modal logics of belief. The extension also includes a distinguished function symbol $pl$ that assigns a numeric value to each situation. The function $pl$ is a *plausibility function*, and the intended interpretation of $pl(s_1) < pl(s_2)$ is that $s_1$ is more plausible than $s_2$. The formula $Bel(\phi, s)$ which expresses the fact that $\phi$ is believed in situation $s$, is defined as follows:

$$Bel(\phi, s) \iff \forall s'[B(s', s) \wedge (\forall s'' B(s'', s) \rightarrow pl(s') \leq pl(s''))] \rightarrow \phi[s'].$$

This formula states that $\phi$ is believed in $s$ if and only if $\phi$ is true in every $B$-accessible situation that is assigned a minimal $pl$-value. In other words, $\phi$ is believed if it is true in all of the most plausible worlds that are considered possible.

Note that the accessibility relation $B$ can be used to define a formula $init(s)$ that defines a set of *initial situations*:

$$init(s) \Leftrightarrow B(s, S_0).$$

The set of situations initially believed possible are the $pl$-minimal elements of $init$. Since $init$ is a formula with one free variable, it defines a set of situations in a given first-order theory. We will let $init^{\mathcal{M}}$ denote the set of situations that satisfy $init$ in the interpretation $\mathcal{M}$. The successor state axiom for $pl$ is straightforward, and it guarantees the following condition:

$$pl(do(a, s)) = pl(s).$$

In order to express the successor state axiom for $B$, it is convenient to distinguish between *ontic actions* that change the state of the world, and *sensing actions* that simply give an agent information about the world. For ontic actions, the successor state axiom for $B$ guarantees the following:

$$B(s', do(A, s)) \iff \exists s''(B(s'', s) \land s' = do(A, s'')).$$

This axiom states that $s$ is accessible after executing $A$ if and only if $s$ is accessible from some world that results from executing $A$ in a state that is considered possible. The effects of binary sensing actions are given through a special sensing predicate $SF$ (Levesque, 1996). For our purposes, it will be sufficient to restrict attention to sensing actions that simply determine the truth value of a single fluent symbol. For sensing actions, the successor state axiom for $B$ says that $s'$ is $B$-related to $do(O, s)$ just in case $s'$ agrees with $s$ on the value of the sensed fluent symbol. The effects of sensing actions in the SitCalc define an approach to belief revision that satisfies five of the AGM postulates (Shapiro et al., 2000).

In order to compare belief change in the SitCalc with belief evolution, we need to express situations in terms of states. In order to simplify the discussion, we will restrict attention to SitCalc action theories where every fluent symbol is unary, with the exception of the distinguished accessibility fluent $B$. We say that a SitCalc theory $\mathcal{T}$ is *elementary* if it satisfies the following conditions:

1. The set of fluent symbols is $\mathbf{F} \cup \{B\}$, where each $F \in \mathbf{F}$ is unary.

2. $\mathcal{T}$ is complete.

3. Every interpretation with $\mathcal{M} \models \mathcal{T}$ has the following properties:

   (a) $\mathcal{M} \models init(S_0)$.

   (b) $|\{x \mid init^{\mathcal{M}}(x)\}| = 2^{|\mathbf{F}|}$.

   (c) If $\mathcal{M} \models init(s_1) \land init(s_2) \land s_1 \neq s_2$, then there is some fluent symbol $F \in \mathbf{F}$ such that $\mathcal{M} \models F(s_i)$ for exactly one of $s_1$ and $s_2$.

Elementary SitCalc theories are essentially categorical on the set of initial situations. Given a model $\mathcal{M}$ of an elementary SitCalc theory, each situation $s^{\mathcal{M}}$ defines a propositional interpretation $I_s^{\mathcal{M}}$ over the set $\mathbf{F}$ of unary fluent symbols. Specifically, for each situation $s^{\mathcal{M}}$, we define $I_s^{\mathcal{M}}$ as follows:

$$I_s^{\mathcal{M}} \models F \iff F^{\mathcal{M}}(s^{\mathcal{M}}).$$

We can also use this idea to associate a *belief state* with every situation $s^{\mathcal{M}}$. For ease of readability, in the following definition we omit the superscript $\mathcal{M}$ on all situation terms and fluent symbols on the right hand side:

$$\kappa_{(s,\mathcal{M})} = \bigcup \{ I_{s'} \mid B(s',s) \text{ and } (\forall s'' B(s'',s) \to pl(s') \leq pl(s'')) \}.$$

Hence $\kappa_{(s,\mathcal{M})}$ is the set of states with minimal plausibility among the situations that are are $B^{\mathcal{M}}$-accessible from $s^{\mathcal{M}}$.

For a model $\mathcal{M}$ of an elementary SitCalc theory $\mathcal{T}$, we define belief update, belief revision and belief evolution. For any ontic action symbol $A$, define $\diamond_{\mathcal{M}}$ as follows.

$$\kappa_{(s,\mathcal{M})} \diamond_{\mathcal{M}} A = \kappa_{(do(A,s),\mathcal{M})}.$$

Note that the plausibility function $pl^{\mathcal{M}}$ defines a total pre-order over the initial situations. Since the initial situations are in 1-1 correspondence with interpretations of $\mathbf{F}$, it follows that $pl^{\mathcal{M}}$ defines a total pre-order over interpretations of $\mathbf{F}$. Let $*_{\mathcal{M}}$ denote the AGM revision operator corresponding to this ordering. Finally, let $\circ_{\mathcal{M}}$ denote the belief evolution operator obtained from $\diamond_{\mathcal{M}}$ and $*_{\mathcal{M}}$. For ease of readability, we will omit the subscript when the theory $\mathcal{M}$ is clear.

In order to state our main result concisely, we introduce some simplifying notation. Let $O$ be a sensing action symbol for the fluent symbol $F_O$, and let $\mathcal{M}$ be a suitable first-order interpretation. We define a set of *propositional interpretations* over $\mathbf{F}$ as follows:

$$O_s^{\mathcal{M}} = \begin{cases} \{ I \mid I \models F_0 \}, & \text{if } F_O^{\mathcal{M}}(s^{\mathcal{M}}) \\ \{ I \mid I \models \neg F_0 \}, & \text{otherwise} \end{cases}$$

Inuitively $O_s^{\mathcal{M}}$ is the set of interpretations that agree with $I_s^{\mathcal{M}}$ on the value of the fluent $F_O$. In the following result, let $\kappa_{(s,\mathcal{M})} \circ \langle A, \alpha \rangle$ stand for the final belief state given by this belief evolution operation.

**Proposition 10** *Let $\mathcal{T}$ be an elementary SitCalc theory, let $\mathcal{M} \models \mathcal{T}$ and let $\circ$ be the evolution operator induced by $\mathcal{M}$. If $O$ is a sensing action and $A$ is an ontic action, then:*

$$\kappa_{(do([A,O],S_0),\mathcal{M})} = \kappa_{(S_0,\mathcal{M})} \circ \langle A, O_{S_0}^{\mathcal{M}} \rangle.$$

**Proof** Without loss of generality, assume that $\mathcal{M} \models F_O(do(A,S_0))$. In other words, assume that $F_O$ holds in the situation resulting from executing the action $A$. Under this assumption, we must prove

$$\kappa_{(do([A,O],S_0),\mathcal{M})} = \kappa_{(S_0,\mathcal{M})} * |F_O|^{-1}(A) \diamond A.$$

If $F_O$ happens to be false after executing $A$, the changes required in the proof are obvious. Note that $\mathcal{M} \models B(s, do([A,O],S_0))$ just in case:

1. $s = do(A, s_1)$ for some $s_1$ with $\mathcal{B}^{\mathcal{M}}(s_1^{\mathcal{M}}, S_0^{\mathcal{M}})$, and

2. $F_O^{\mathcal{M}}(s)$.

Suppose that $I \in \kappa_{(do([A,O],S_0),\mathcal{M})}$. It follows that $I = I_s^{\mathcal{M}}$ for some situation term $s$ that satisfies conditions (1) and (2), and also has the property that $s^{\mathcal{M}}$ is $pl^{\mathcal{M}}$-minimal among situations that are $B^{\mathcal{M}}$ accessible from $do([A,O], S_0)^{\mathcal{M}}$ . Consider the situation term $s_1$ in condition (1). Clearly $I_{s_1}^{\mathcal{M}} \in |F_O|^{-1}(A)$, because $\mathcal{M} \models F_O(do(A, s'))$. Now suppose there exists $s_2$ with the following properties:

1. $I_{s_2}^{\mathcal{M}} \in |F_O|^{-1}(A)$

2. $\mathcal{M} \models B(do([A,O], s_2), do([A,O]), S_0))$

3. $pl^{\mathcal{M}}(s_2^{\mathcal{M}}) < pl^{\mathcal{M}}(s_1^{\mathcal{M}})$.

By the successor state axiom for $pl$, it follows that

$$pl^{\mathcal{M}}(do(A, s_2)^{\mathcal{M}}) < pl^{\mathcal{M}}(do(A, s_1)^{\mathcal{M}}) = pl^{\mathcal{M}}(s^{\mathcal{M}}).$$

This contradicts the $pl^{\mathcal{M}}$-minimality of $s^{\mathcal{M}}$ among situations that are $B^{\mathcal{M}}$ accessible from $do([A,O], S_0)^{\mathcal{M}}$. Therefore, there is no such $s_2$. So $s_1^{\mathcal{M}}$ is $pl^{\mathcal{M}}$-minimal among situations that satisfy the first two properties above defining $s_2$.

Recall that $\mathcal{M}$ is elementary, so every propositional interpretation $I \in |F_O|^{-1}(A)$ is equal to $I_t^{\mathcal{M}}$ for some initial situation $t$. Note that, if $I \in |F_O|^{-1}(A)$, then $I = I_t^{\mathcal{M}}$ for some $t$ that satisfies points (1) and (2) in the specification of $s_2$ above. If $I \in |F_O|^{-1}(A)$ is less than $I_{s_1}^{\mathcal{M}}$ in the total pre-order on *interpretations* defined by $pl^{\mathcal{M}}$, then this situation $t$ also satisfies the third condition. We have seen that this is not possible. So, if $\leq_{pl}^{\mathcal{M}}$ is the total pre-order on interpretations defined by $pl^{\mathcal{M}}$, then we have:

$$\begin{aligned} I_{s_1}^{\mathcal{M}} &\in& \min(|F_O|^{-1}(A), \leq_{pl}^{\mathcal{M}}) \\ &=& \kappa_{(S_0,\mathcal{M})} * |F_O|^{-1}(A) \qquad \text{(by definition)} \end{aligned}$$

But then, since $s^{\mathcal{M}} = do(A, s_1)^{\mathcal{M}}$:

$$I_s^{\mathcal{M}} \in \kappa_{(S_0,\mathcal{M})} * |F_O|^{-1}(A) \diamond A.$$

So $\kappa_{(do([A,O],S_0),\mathcal{M})} \subseteq \kappa_{(S_0,\mathcal{M})} * |F_O|^{-1}(A) \diamond A$.

For the other direction, suppose that $I \in \kappa_{(S_0,\mathcal{M})} * |F_O|^{-1}(A) \diamond A$. So there is some $I' \in \kappa_{(S_0,\mathcal{M})} * |F_O|^{-1}(A)$ such that $I' \diamond A = I$. But then $I' = I_s^{\mathcal{M}}$ for some $pl^{\mathcal{M}}$-minimal situation $s^{\mathcal{M}} \in init^{\mathcal{M}}$ such that $\mathcal{M} \models F_O(do(A, s))$. Since $s^{\mathcal{M}} \in init^{\mathcal{M}}$, it follows that

$$\mathcal{M} \models B(do(A, s), do(A, S_0)).$$

Moreover, since $\mathcal{M} \models F_O(do(A, s))$, it follows that

$$\mathcal{M} \models B(do([A,O], s), do([A,O], S_0)).$$

Now suppose that there is a situation term $s_1$ such that $\mathcal{M} \models B(s_1, do([A,O], S_0))$ and $pl^{\mathcal{M}}(s_1^{\mathcal{M}}) < pl^{\mathcal{M}}(do(A, s))^{\mathcal{M}}$. It follows immediately that there is some $s_2^{\mathcal{M}} \in init^{\mathcal{M}}$ such that $do([A, O], s_2)^{\mathcal{M}} = s_1^{\mathcal{M}}$. Moreover, by the successor state axiom for $pl$, we have $pl^{\mathcal{M}}(s_2^{\mathcal{M}}) < pl^{\mathcal{M}}(s^{\mathcal{M}})$. However, since $s_2^{\mathcal{M}} \in init^{\mathcal{M}}$, this contradicts the $pl^{\mathcal{M}}$ minimality of $s^{\mathcal{M}}$ among initial situations. So there is no such $s_1^{\mathcal{M}}$. Therefore $I_{do(A,s)} \in \kappa_{do([A,O],S_0)}$. Recall that $I_{do(A,s)}^{\mathcal{M}} = I_s^{\mathcal{M}} \diamond A = I$. Therefore, $I \in \kappa_{(do([A,O],S_0),\mathcal{M})}$. $\quad\square$

Skimming over the details, the preceding proof simply relies on the condition that $\mathcal{M} \models pl(do(a, s)) = pl(s)$ in every epistemic SitCalc model. The fact that plausibility values persist following the execution of actions is equivalent to restricting belief change by always revising the initial belief state, then determining the final belief state by simply computing ontic action effects. Hence, the semantics of revision actions in the SitCalc can be framed as an instance of belief evolution. We suggest that this is an important point in discussing belief change in the SitCalc. In the original description of the approach, the belief change due to a sensing action is identified with belief revision (Shapiro et al., 2000). On this view, the fact that sensing actions do not satisfy all of the AGM postulates can be seen as a problem with the approach. However, when we are explicit about the fact that belief change in the SitCalc is a form of belief evolution, this is no longer a problem. We do not expect the AGM postulates to be satisfied, as we need to be concerned with the interaction between ontic actions and sensing actions.

We conclude this section by remarking that belief evolution operators do have one expressive advantage over the epistemic extension of the SitCalc. In the epistemic extension that we have considered, the information obtained from sensing actions is always correct. As a result, it is not sensible to consider revising by $F$ followed by $\neg F$. By contrast, in belief evolution, this is simply handled by keeping the more reliable observation. Hence, belief evolution is able to deal with unreliable perception in a straightforward manner that is not possible in the SitCalc. We remark however, that inconsistent observations have been treated in a later extension of the SitCalc postulating exogenous actions to account for the inconsistent sensing information (Shapiro & Pagnucco, 2004).

## 6.2 Iterated Belief Revision

Recall that $\lambda$ is a null action that does not change the state of the world. If we consider action domains where agents can perform $\lambda$, then sequential belief revision is a special case of belief evolution.

**Observation 1** *For any $\kappa$ and $\bar{\alpha}$, there is a unique belief state $\kappa'$ such that*

$$\kappa \circ \langle \bar{\lambda}, \bar{\alpha} \rangle = \langle \kappa', \ldots, \kappa' \rangle.$$

Since every action is null, the unique belief state $\kappa'$ is the belief state that results after the sequence $\alpha$ of observations. In this section, we consider belief evolution operators from the perspective of the well-known Darwiche-Pearl postulates for iterated revision (Darwiche & Pearl, 1997).

First, it is important to note that belief evolution does not define iterated revision as a simple sequence of AGM revision operations. According to Definition 11, inconsistencies between observations are resolved by keeping the more reliable observations. By default, we take recency as our measure of reliability. To illustrate, consider a simple example. In the following expression, let $\tilde{\alpha}$ denote the complement of $\alpha$

$$\kappa \circ \langle\langle \lambda, \lambda\rangle, \langle \tilde{\alpha}, \alpha\rangle\rangle.$$

In this example, the observation $\tilde{\alpha}$ is followed by the observation $\alpha$. The final belief state is not obtained by performing two single-shot revisions, however. According to Definition 11 with the recency ordering, the first observation is discarded because it is inconsistent with a more recent observation. As such, the final belief state after this operation is $\kappa * \alpha$. Hence, the approach to iterated revision that is implicit in belief evolution is not a naive iteration. It is therefore reasonable to ask if the implicit iterated revision operator satisfies existing rationality postulates.

We state the Darwiche-Pearl postulates in terms of possible worlds. Let $\kappa$, $\alpha$, and $\beta$ be sets of possible worlds. The Darwiche-Pearl postulates are as follows.

**Darwiche-Pearl Postulates**
[DP1] If $\alpha \subseteq \beta$, then $(\kappa * \beta) * \alpha = \kappa * \alpha$.
[DP2] If $\alpha \subseteq \tilde{\beta}$, then $(\kappa * \beta) * \alpha = \kappa * \alpha$.
[DP3] If $\kappa * \alpha \subseteq \beta$, then $(\kappa * \beta) * \alpha \subseteq \beta$.
[DP4] If $\kappa * \alpha \not\subseteq \tilde{\beta}$, then $(\kappa * \beta) * \alpha \not\subseteq \tilde{\beta}$.

We would like to determine if these postulates hold when we perform belief evolution with null actions.

We *define* the iterated revision operator *obtained from* $\circ$ as follows:

$$\kappa * \beta * \alpha =_{def} \kappa \circ \langle \bar{\lambda}, \langle \beta, \alpha\rangle\rangle. \tag{6}$$

Of course this is not true if we apply the AGM operator on the left successively. But we adopt this convention as a definition of iterated revision under $\circ$ because it allows us to ask if the Darwiche-Pearl postulates hold. It is critical to remark, however, that we are using a notational convention that defines iterated revision in terms of belief evolution.

Given that we have to introduce a new notational convention to even ask if the Darwiche-Pearl postulates hold, one might question why we are concerned with these postulates. In other words, why is it important to check that belief evolution operators satisfy key properties of *iterated belief revision*? Our stance is that these postulates matter for belief evolution because we can en up using belief evolution operators for iterated revision "by accident." Although our focus is on iterated sequences of actions and observations, there will be cases where the actions are all null and we are clearly looking at a case of iterated revision. It would be problematic if belief evolution handled such cases poorly, so we would like to ensure that these instances of iterated revision are handled appropriately. One way to assess "appropriateness" is by checking if the Darwiche-Pearl postulates hold.

The next result relies on the following crucial observation. If $\alpha \neq \emptyset$, then

$$\kappa * \beta * \alpha = \begin{cases} \kappa * (\beta \cap \alpha) & \text{if } \beta \cap \alpha \neq \emptyset \\ \kappa * \alpha & \text{otherwise} \end{cases}$$

This observation follows immediately from the definition of belief evolution. By using this expression, we can prove the following result.

**Proposition 11** *Let $\circ$ be a belief evolution operator and let $\alpha \neq \emptyset$. Then the iterated revision operator as given in (6) satisfies the Darwiche-Pearl postulates.*

**Proof** Note that $\alpha^{-1}(\lambda) = \alpha$ and $\beta^{-1}(\lambda) = \beta$. Since $\alpha \neq \emptyset$, we need to show that the DP postulates are satisfied by $\kappa * \beta * \alpha$, as defined in the observation above.

For [DP1], suppose that $\alpha \subseteq \beta$. Since $\alpha \neq \emptyset$, it follows that $\beta \cap \alpha \neq \emptyset$ and hence $\kappa * \beta * \alpha = \kappa * (\beta \cap \alpha)$. But $\beta \cap \alpha = \alpha$, so the right hand side is equal to $\kappa * \alpha$.

For [DP2], suppose that $\alpha \subseteq \tilde{\beta}$. Hence, $\alpha \cap \beta = \emptyset$ and the desired conclusion follows immediately.

For [DP3], suppose that $\kappa * \alpha \subseteq \beta$. Since $\alpha \neq \emptyset$, it follows from [AGM5] that $\kappa * \alpha \neq \emptyset$. So there exists some $s \in \kappa * \alpha$. But then $s \in \alpha$ since $\kappa * \alpha \subseteq \alpha$, and $s \in \beta$ since $\kappa * \alpha \subseteq \beta$. Hence $\alpha \cap \beta \neq \emptyset$, and therefore

$$\kappa * \beta * \alpha = \kappa * (\beta \cap \alpha) \subseteq \beta \cap \alpha \subseteq \beta.$$

For [DP4], suppose that $\kappa * \alpha \not\subseteq \tilde{\beta}$. So there exists some $s \in \kappa * \alpha$ such that $s \in \beta$. It follows that $s \in \alpha \cap \beta$, so $\alpha \cap \beta \neq \emptyset$. Translating to possible worlds, postulate [AGM7] says the following:

$$\text{if } \kappa * \alpha \not\subseteq \tilde{\beta}, \text{ then } (\kappa * \alpha) \cap \beta \subseteq \kappa * (\alpha \cap \beta).$$

Since $s \in (\kappa * \alpha) \cap \beta$, this implies that $s \in \kappa * (\alpha \cap \beta)$. But then, since $\alpha \cap \beta \neq \emptyset$ it follows by definition that $s \in \kappa * \beta * \alpha$. Hence $s \in \beta$ and $s \in \kappa * \beta * \alpha$. Therefore $s \in \kappa * \beta * \alpha \not\subseteq \tilde{\beta}$. $\square$

It is well known that many AGM revision operators do not satisfy the Darwiche-Pearl postulates when applied in succession. Proposition 11 shows that, even if we start with a single-shot revision operator, the approach to iterated revision induced by belief evolution is always a Darwiche-Pearl operator.

It is easy to demonstrate that belief evolution also satisfies the so-called *recalcitrance* postulate introduced by Nayak, Pagnucco and Peppas (2003). Rephrased in terms of possible worlds and belief evolution, recalcitrance is the following property:

(**Recalcitrance**) If $\beta \cap \alpha \neq \emptyset$, then $(\kappa \circ \langle \bar{\lambda}, \langle \beta, \alpha \rangle \rangle) \subseteq \beta$.

It is known that DP1, DP2, and (Recalcitrance) characterize Nayak's lexicographic iterated revision operator on epistemic states (Booth & Meyer, 2006). It follows from this that our approach to iterated revision also satisfies the independence postulate introduced independently by Jin and Thielscher (2007) as well as Booth and Meyer (2006). This gives a complete characterization of the iterated revision operator that is implicit in belief evolution, from the perspective of the Darwiche-Pearl tradition.

### 6.3 Lehmann Postulates

In this section, we consider belief evolution from the perspective of Lehmann's postulates (1995). Given observation trajectories $O$ and $O'$, let $O \cdot O'$ denote the concatenation of the two sequences. If $\alpha$ is an observation, we write $O \cdot \alpha$ as a shorthand for $O \cdot \langle \alpha \rangle$. Finally, for the remainder of this section we write $\kappa \circ O$ as an abbreviation for the final belief state in $\kappa \circ \langle \bar{\lambda}, O \rangle$. Translated into our notation, the Lehmann postulates are as follows.

**Lehmann Postulates**

[L2] $\kappa \circ (O \cdot \alpha) \subseteq \alpha$.

[L3] If $\kappa \circ (O \cdot \alpha) \subseteq \beta$ and $\kappa \circ O \subseteq \alpha$, then $\kappa \circ O \subseteq \beta$.

[L4] If $\kappa \circ O \subseteq \alpha$, then $\kappa \circ (O \cdot O') = \kappa \circ (O \cdot \alpha \cdot O')$.

[L5] If $\beta \subseteq \alpha$, then $\kappa \circ (O \cdot \alpha \cdot \beta \cdot O') = \kappa \circ (O \cdot \beta \cdot O')$.

[L6] If $\kappa \circ (O \cdot \alpha) \not\subseteq \tilde{\beta}$, then $\kappa \circ (O \cdot \alpha \cdot \beta \cdot O') = \kappa \circ (O \cdot \alpha \cdot \alpha \cap \beta \cdot O')$.

[L7] $\kappa \circ (O \cdot \alpha) \subseteq \kappa \circ (O \cdot \tilde{\alpha} \cdot \alpha)$.

We start with postulate [L2] to remain consistent with the original numbering. However, we omit Lehmann's first postulate as it states that sequences of observed formulas define a consistent theory. Since we work directly with sets of states rather than formulas, this kind of postulate is not necessary.

From the perspective of iterated belief change, a more interesting distinction between our approach and Lehmann's approach can be seen by looking at postulates [L4]-[L6]. Lehmann views [L4]-[L6] as dealing with "superfluous revisions" (Lehmann, 1995). For example, in postulate [L4], the observation $\alpha$ is superfluous because revising by the observations in $O$ already leads an agent to believe that the actual state is in $\alpha$. As such, observing $\alpha$ after $O$ does not provide any new information. The postulate [L4] suggests that such observations may be discarded. This kind of reasoning is not supported in belief evolution, because the observation $\alpha$ may take on new meaning following future observations. Postulates [L5] and [L6] are problematic for similar reasons.

We present a counterexample that illustrates that postulates [L4]-[L6] all fail for belief evolution.

**Example** Let $s_1, s_2, s_3$ be states over some action signature. Define $\kappa$, $\alpha$, $O$ and $O'$ as follows:

$$\begin{aligned}
\kappa &= \{s_1\} \\
\alpha &= \{s_2, s_3\} \\
\beta &= \{s_3\} \\
O &= \langle \{s_3\} \rangle \\
O' &= \langle \{s_1, s_2\} \rangle
\end{aligned}$$

We demonstrate that [L4]-[L6] all fail for this example.

Let $\circ$ be a belief evolution operator obtained from some update operator $\diamond$ and some AGM revision operator $*$. Note that

$$\kappa \circ O = \{s_1\} * \{s_3\} = \{s_3\} \subseteq \alpha.$$

However,

$$\begin{aligned}
\kappa \circ (O \cdot O') &= \{s_1\} * \{s_1, s_2\} = \{s_1\} \\
\kappa \circ (O \cdot \alpha \cdot O') &= \{s_1\} * \{s_2\} = \{s_2\}.
\end{aligned}$$

Hence $\kappa \circ (O \cdot O') \neq \kappa \circ (O \cdot \alpha \cdot O')$, which violates [L4]. Since $\beta = \kappa \circ O$, this also violates [L5].

Let $\gamma = \{s_1, s_3\}$. The following equalities refute [L6].

$$
\begin{aligned}
\kappa \circ (O \cdot \alpha) &= \{s_3\} \not\subseteq \tilde{\gamma} \\
\kappa \circ (O \cdot \alpha \cdot \gamma \cdot O') &= \{s_1\} \\
\kappa \circ (O \cdot \alpha \cdot \alpha \cap \gamma \cdot O') &= \{s_2\}.
\end{aligned}
$$

The preceding example demonstrates that [L4]-[L6] do not hold for belief evolution; however, it is perhaps too abstract to illustrate the intuitive problem. At the level of commonsense reasoning, it is instructive to imagine a situation involving a certain bird that is either red, yellow or black. Postulate [L4] says that the following informal sequences lead to the same belief state:

Believe(*bird is red*) + Observe(*black, red OR yellow*)
Believe(*bird is red*) + Observe(*black, yellow OR black, red OR yellow*)

However, we can see that such sequences are not the same from the perspective of our belief evolution operators. In both sequences, the first observation is discarded[2]. In the first case, the agent is left to keep the initial belief that the bird is red because that is consistent with the final observation. By contrast, in the second case, the final two observations are combined to suggest that the bird is yellow. Following the intuitions of AGM revision, the agent then believes the bird is yellow. Similar "bird colour" arguments can be used to demonstrate whey [L5] and [L6] fail for belief evolution.

Although [L4]-[L6] do not hold, we can construct weaker versions that do hold. We have claimed that the reason these postulates fail is because future observations may affect the interpretation of observations that are initially superfluous. To avoid this problem, we modify the postulates by removing the observations that follow a superfluous observation. Weakening gives the following postulates:

**Weak Lehmann Postulates**
[L4*] If $\kappa \circ O \subseteq \alpha$ and $O \neq \tilde{\emptyset}$, then $\kappa \circ O = \kappa \circ (O \cdot \alpha)$.
[L5*] If $\beta \subseteq \alpha$, then $\kappa \circ (O \cdot \alpha \cdot \beta) = \kappa \circ (O \cdot \beta)$.
[L6*] If $\kappa \circ (O \cdot \alpha) \not\subseteq \tilde{\beta}$, then $\kappa \circ (O \cdot \alpha \cdot \beta) = \kappa \circ (O \cdot \alpha \cdot \alpha \cap \beta)$.

If [L4]-[L6] are replaced with [L4*]-[L6*], then belief evolution satisfies the resulting set of postulates.

**Proposition 12** *Let $\circ$ be a belief evolution operator, let $\alpha \neq \emptyset$ and let $O$ be an observation trajectory where each component is non-empty. Then $\circ$ satisfies [L2], [L3], [L4*], [L5*], [L6*], and [L7].*

---

2. The first observation is only discarded under the assumption that *recent* information takes precedence. As discussed previously, we are not committed to this assumption, but it is useful for the purpose of comparison with iterated belief revision.

**Proof** Since $\alpha \neq \emptyset$, $\kappa \circ (O \cdot \alpha) = \kappa * \gamma$ for some $\gamma \subseteq \alpha$. Hence $\kappa \circ (O \cdot \alpha) \subseteq \gamma \subseteq \alpha$, which proves [L2].

For [L3], suppose that $\kappa \circ (O \cdot \alpha) \subseteq \beta$ and $\kappa \circ O \subseteq \alpha$. Now suppose that $s \in \kappa \circ O$. By definition, this means that $s \in \kappa * \bigcap \tau(O)$. By postulate [AGM7], we have $(\kappa * \bigcap \tau(O)) \cap \alpha \subseteq \kappa * (\bigcap \tau(O) \cap \alpha)$. But $\kappa * (\bigcap \tau(O) \cap \alpha) = \kappa \circ (O \cdot \alpha)$, so it follows that $s \in \beta$. Therefore $\kappa \circ O \subseteq \beta$.

Suppose that $\kappa \circ O \subseteq \alpha$ and $O \neq \bar{\emptyset}$. By definition, $\kappa \circ O = \kappa * \bigcap \tau(O)$. Since $O \neq \bar{\emptyset}$, it follows that $\kappa \circ (O \cdot \alpha) = \kappa * (\bigcap \tau(O) \cap \alpha)$. With these equalities in mind, the following results prove that [L4*] holds:

$$
\begin{aligned}
\kappa * \bigcap \tau(O) &\subseteq \quad (\kappa * \bigcap \tau(O)) \cap \alpha \quad &&(\text{since } \kappa \circ O \subseteq \alpha) \\
&\subseteq \quad \kappa * (\bigcap \tau(O) \cap \alpha) \quad &&(\text{by [AGM7]}) \\
&\subseteq \quad \kappa * \bigcap \tau(O) \quad &&(\text{by [AGM8]})
\end{aligned}
$$

It is clear that [L5*] holds, simply because the assumption that $\beta \subseteq \alpha$ implies that $\beta = \alpha \cap \beta$.

For [L6*], suppose that $\kappa \circ (O \cdot \alpha) \not\subseteq \tilde{\beta}$. It follows that $\alpha \cap \beta \neq \emptyset$. By definition, this means that $\kappa \circ (O \cdot \alpha \cdot \beta) = \kappa \circ (O \cdot \alpha \cdot \alpha \cap \beta)$, which is the desired result.

Since $\tilde{\alpha} \cap \alpha = \emptyset$, it follows by definition that $\kappa \circ (O \cdot \alpha) = \kappa \circ (O \cdot \tilde{\alpha} \cdot \alpha)$. □

Hence, if we do not consider the influence of observations that will occur in the future, then belief evolution defines an approach to iterated revision that satisfies all of the Lehmann postulates that do not deal with empty belief states.

We conclude with a brief remark about the assumption that $\alpha$ and $\beta$ are non-empty in Propositions 11 and 12. In our framework, action histories take precedence over observations. As such, empty observations are discarded because they are inconsistent with every sequence of actions. This is true even in the case when $\bar{A}$ is a sequence of null actions. We accept this treatment of inconsistent observations, because it allows inconsistency to be treated in a uniform manner.

## 7. Discussion

We have presented a transition system framework for reasoning about belief change due to actions and observations. We suggested that agents should perform belief update following an action, and they should perform belief revision following an observation. We showed that the interaction between update and revision can be non-elementary. Consequently we specified how an agent should consider the history of actions when incorporating a new observation. In contrast, existing formalisms for reasoning about epistemic action effects either ignore the interaction between revision and update or they deal with it implicitly. Hence, we have provided an explicit treatment of a phenomenon that has not always been recognised in related formalisms.

The fundamental idea motivating this work is that the interpretation of an observation may depend on the preceding sequence of actions. Our treatment of iterated belief change includes two main components. First, we introduce a set of postulates that capture our intuitions about the appropriate treatment of an observation following a sequence of actions.

Informally, our postulates capture the intuitions of AGM revision for domains where actions may occur. Second, we introduce belief evolution operators to give a concrete recipe for combining a given update operator with a given revision operator. Belief evolution operators satisfy our postulates, along with some standard postulates for iterated revision.

The class of problems that are appropriate for belief evolution can be described by an ordering over action histories. Let $A_1, \alpha_1, \ldots, A_n, \alpha_n$ be an alternating sequence of actions and observations. Let $\preceq$ denote a total pre-order over the elements of this sequence. Belief evolution is suitable for problems in which the underlying ordering is given as follows, for some permutation $p_1, \ldots, p_n$ of $1, \ldots, n$.

$$\left.\begin{array}{c} A_1 \\ \vdots \\ A_n \end{array}\right\} \preceq \alpha_{p_1} \preceq \alpha_{p_2} \preceq \cdots \preceq \alpha_{p_n}$$

Hence belief evolution is appropriate when we have any total pre-order over observations, including the case where all observations are considered equally reliable. However, we have not addressed the case where action histories may be incorrect.

We have framed our results in a transition system framework, primarily because it provides a simple representation of action effects. However, our results can be translated into related action formalisms as well. For example, in our comparison with the SitCalc, we illustrated that the revision actions of the SitCalc are implicitly defined in terms of belief evolution. Our results provide a general account of the interaction between actions and observations, grounded in the context of transition systems.

It is worth noting that belief evolution operators model a specific reasoning problem that has been addressed in more general formalisms. For example, in our own work, we have explored the use of arbitrary plausibility rankings over both actions and observations to reason about iterated belief change (Hunter & Delgrande, 2006). Moreover, even the distinction between revision and update can be understood as a pragmatic distinction for modelling certain kinds of reasoning. A single generic notion of "belief change" has been defined such that belief revision and belief update are both special cases (Kern-Isberner, 2008). From this perspective, iterated belief change due to action is simply one particular instance; it is an instance where the belief change operations are subject to certain constraints that encode the effects of actions. We believe this is an important instance that is worthy of detailed study, but it is important to be aware that it can be framed and understood at a more general level.

There are several directions for future research. One direction deals with the implementation of a belief evolution solver. We have previously explored the use of Answer Set Planning to develop a solver for iterated belief change (Hunter, Delgrande, & Faber, 2007), and we believe this approach could be further developed. Another important direction for future research involves relaxing the assumption that action histories are correct. In realistic action domains, agents may be incorrect about the actions that have occurred. In such domains, it is not always reasonable to discard an observation that is inconsistent with the perceived action history. Instead, an agent should consider the likelihood that the observation is correct as well as the likelihood that the action history is correct. Hence, the most plausible histories can be determined by considering the relative plausibility of each action and observation. Belief evolution operators can be seen as a specific case of this kind of reasoning, in which action occurrences are always more plausible than observations.

# References

Alchourrón, C., Gärdenfors, P., & Makinson, D. (1985). On the logic of theory change: Partial meet functions for contraction and revision. *Journal of Symbolic Logic*, *50*(2), 510–530.

Booth, R., & Meyer, T. (2006). Admissible and restrained revision. *Journal of Artificial Intelligence Research*, *26*, 127–151.

Boutilier, C. (1995). Generalized update: Belief change in dynamic settings. In *Proceedings of the Fourteenth International Joint Conference on Artificial Intelligence (IJCAI 1995)*, pp. 1550–1556.

Darwiche, A., & Pearl, J. (1997). On the logic of iterated belief revision. *Artificial Intelligence*, *89*(1-2), 1–29.

Delgrande, J., Dubois, D., & Lang, J. (2006). Iterated revision as prioritized merging. In *Proceedings of the 10th International Conference on Principles of Knowledge Representation and Reasoning (KR2006)*.

Gelfond, M., & Lifschitz, V. (1998). Action languages. *Linköping Electronic Articles in Computer and Information Science*, *3*(16), 1–16.

Herzig, A., Lang, J., & Marquis, P. (2004). Revision and update in multi-agent belief structures. In *Proceedings of LOFT 6*.

Hunter, A., & Delgrande, J. (2005). Iterated belief change: A transition system approach. In *Proceedings of the International Joint Conference on Artificial Intelligence (IJCAI05)*, pp. 460–465.

Hunter, A., & Delgrande, J. (2006). Belief change in the context of fallible actions and observations. In *Proceedings of the National Conference on Artificial Intelligence(AAAI06)*.

Hunter, A., Delgrande, J., & Faber, J. (2007). Using answer sets to solve belief change problems. In *Proceedings of the 9th International Conference on Logic Programming and Non Monotonic Reasoning (LPNMR 2007)*.

Jin, Y., & Thielscher, M. (2007). Iterated belief revision, revised. *Artificial Intelligence*, *171*(1), 1–18.

Katsuno, H., & Mendelzon, A. (1991). On the difference between updating a knowledge base and revising it. In *Proceedings of the Second International Conference on Principles of Knowledge Representation and Reasoning (KR 1991)*, pp. 387–394.

Katsuno, H., & Mendelzon, A. (1992). Propositional knowledge base revision and minimal change. *Artificial Intelligence*, *52*(2), 263–294.

Kern-Isberner, G. (2008). Linking iterated belief change operations to nonmonotonic reasoning. In *Proceedings of the 11th International Conference on Principles of Knowledge Representation and Reasoning (KR2008)*.

Lang, J. (2006). About time, revision, and update. In *Proceedings of the 11th International Workshop on Non-Monotonic Reasoning (NMR 2006)*.

Lehmann, D. (1995). Belief revision, revised. In *Proceedings of the Fourteenth International Joint Conference on Artificial Intelligence (IJCAI95)*, pp. 1534–1541.

Levesque, H. (1996). What is planning in the presence of sensing?. In *Proceedings of the Thirteenth National Conference on Artificial Intelligence (AAAI96)*, pp. 1139–1146.

Levesque, H., Pirri, F., & Reiter, R. (1998). Foundations for the situation calculus. *Linköping Electronic Articles in Computer and Information Science*, *3*(18), 1–18.

Lobo, J., Mendez, G., & Taylor, S. (2001). Knowledge and the action description language $\mathcal{A}$. *Theory and Practice of Logic Programming*, *1*(2), 129–184.

Moore, R. (1985). A formal theory of knowledge and action. In Hobbs, J., & Moore, R. (Eds.), *Formal Theories of the Commonsense World*, pp. 319–358. Ablex Publishing.

Nayak, A. (1994). Iterated belief change based on epistemic entrenchment. *Erkenntnis*, *41*, 353–390.

Nayak, A., Pagnucco, M., & Peppas, P. (2003). Dynamic belief change operators. *Artificial Intelligence*, *146*, 193–228.

Papini, O. (2001). Iterated revision operations stemming from the history of an agent's observations. In Rott, H., & Williams, M. (Eds.), *Frontiers in Belief Revision*, pp. 279–301. Kluwer Academic Publishers.

Peppas, P., Nayak, A., Pagnucco, M., Foo, N., Kwok, R., & Prokopenko, M. (1996). Revision vs. update: Taking a closer look. In *Proceedings of the Twelfth European Conference on Artificial Intelligence (ECAI96)*, pp. 95–99.

Shapiro, S., & Pagnucco, M. (2004). Iterated belief change and exogenous actions in the situation calculus. In *Proceedings of the Sixteenth European Conference on Artificial Intelligence (ECAI'04)*, pp. 878–882.

Shapiro, S., Pagnucco, M., Lesperance, Y., & Levesque, H. (2000). Iterated belief change in the situation calculus. In *Proceedings of the Seventh International Conference on Principles of Knowledge Representation and Reasoning (KR 2000)*, pp. 527–538. Morgan Kaufmann Publishers.

Son, T., & Baral, C. (2001). Formalizing sensing actions: A transition function based approach. *Artificial Intelligence*, *125*(1-2), 19–91.